\documentclass[screen, authorversion, nonacm]{acmart}

\usepackage{multirow}
\usepackage{makecell}
\usepackage{multicol}

\usepackage{physics} 
\usepackage{bm}
\usepackage{mathtools} 

\usepackage{algorithm,algcompatible,amsmath}
\renewcommand{\COMMENT}[2][.5\linewidth]{%
  \leavevmode\hfill\makebox[#1][l]{//~#2}}
\algnewcommand\algorithmicto{\textbf{to}}
\algnewcommand\RETURN{\State \textbf{return} }

\usepackage{tikz}
\definecolor{drawioblue}{RGB}{108,142,191}
\definecolor{drawiored}{HTML}{B85450} 
\definecolor{drawioyellow}{HTML}{D6B656} 

\usepackage{xcolor}
\definecolor{lightergray}{gray}{0.9}
\usepackage{mdframed}
\mdfsetup{skipabove=10pt,skipbelow=10pt}

\usepackage{needspace}

\AtBeginDocument{%
  }

\begin{document}

\title{Predictive Coding Networks and Inference Learning: Tutorial and Survey}


\author{Björn van Zwol}
\email{zwolbe@leidenuniv.nl}
\orcid{0009-0003-6542-7064}
\affiliation{
  \institution{Utrecht University}
  \streetaddress{Princetonplein 5}
  \city{Utrecht}
  \country{The Netherlands}
  \postcode{3584 CC}
}
\affiliation{
  \institution{Leiden University}
  \city{Leiden}
  \country{The Netherlands}
  \postcode{2333 CA Leiden}
}

\author{Ro Jefferson}
\email{r.jefferson@uu.nl}
\orcid{0000-0002-2603-0350}
\author{Egon L. van den Broek}
\email{vandenbroek@acm.org}
\orcid{0000-0002-2017-0141}
\affiliation{
  \institution{Utrecht University}
  \streetaddress{Princetonplein 5}
  \city{Utrecht}
  \country{The Netherlands}
  \postcode{3584 CC}
}

\renewcommand{\shortauthors}{van Zwol, et al.}

\begin{abstract}
Recent years have witnessed a growing call for renewed emphasis on neuroscience-inspired approaches in artificial intelligence research, under the banner of \emph{NeuroAI}. A prime example of this is predictive coding networks (PCNs), based on the neuroscientific framework of predictive coding. This framework views the brain as a hierarchical Bayesian inference model that minimizes prediction errors through feedback connections. Unlike traditional neural networks trained with backpropagation (BP), PCNs utilize inference learning (IL), a more biologically plausible algorithm that explains patterns of neural activity that BP cannot. Historically, IL has been more computationally intensive, but recent advancements have demonstrated that it can achieve higher efficiency than BP with sufficient parallelization. Furthermore, PCNs can be mathematically considered a superset of traditional feedforward neural networks (FNNs), significantly extending the range of trainable architectures. As inherently probabilistic (graphical) latent variable models, PCNs provide a versatile framework for both supervised learning and unsupervised (generative) modeling that goes beyond traditional artificial neural networks. This work provides a comprehensive review and detailed formal specification of PCNs, particularly situating them within the context of modern ML methods. This positions PC as a promising framework for future ML innovations.
\end{abstract}
\maketitle


\section{Introduction}\label{sec:intro}
Neuroscience-inspired approaches to machine learning (ML) have a long history within artificial intelligence research \cite{rosenblatt_perceptron_1958, hassabis_neuroscience-inspired_2017, macpherson_natural_2021}.
 Despite the remarkable empirical advances in ML capabilities in recent years, biological learning is still superior in many ways, such as flexibility and energy efficiency \cite{macpherson_natural_2021}.
 As such, recent years have seen a growing call for renewed emphasis on neuroscience-inspired approaches in AI research, known as \textit{NeuroAI} \cite{zador_catalyzing_2023, noauthor_new_2024}. 
This is exemplified by the rising popularity of the predictive coding (PC, also known as predictive processing) framework in computational neuroscience \cite{rao_predictive_1999, friston_free-energy_2010, ficco_disentangling_2021, rao_sensorymotor_2024} and its recent entrance into the field of machine learning \cite{whittington_approximation_2017, <millidge_predictive_towards>}. PC represents a Bayesian perspective on how the brain processes information, emphasizing the role of probabilistic models and minimization of prediction errors in perception and learning \cite{rao_predictive_2002, friston_free-energy_2010, ficco_disentangling_2021, jiang_predictive_2022}. At its core, it suggests that the brain continually generates predictions about sensory input and updates these predictions based on incoming sensory data. By removing the predictable components, this reduces redundancy in the information processing pipeline \cite{huang_predictive_2011}.

\begin{figure}[ ]
  \centering
  \includegraphics[width=0.85\textwidth]{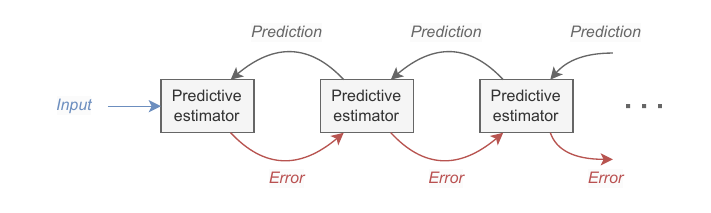}
  \caption{\textbf{Basic PC framework}, in which the error (i.e., the discrepancy) between the actual input from lower layers and the predicted value is the information propagated upwards through the network. Figure adapted from \cite{rao_predictive_1999}.}
  \label{fig:PC_original}
\end{figure}

Central to PC is the concept of \textit{hierarchical prediction}: the lowest level of the hierarchy represents sensory data, and each  higher level attempts to predict neural activity in the layer below \cite{rao_dynamic_1997, rao_predictive_1999}. 
Prediction errors, arising from discrepancies between actual and predicted activity, are propagated upward through the hierarchy, while the predictions from higher levels are propagated downwards via feedback connections; see fig. \ref{fig:PC_original}. In the influential work by Rao \& Ballard (1999), the authors showed that such a predictive coding network (PCN) can learn statistical regularities in the input and explain several neural responses in the visual cortex \cite{rao_development_1998, rao_optimal_1999, rao_predictive_1999}. Since then, the PC framework has found wide adoption across neuroscience, psychology, and philosophy \cite{okada_neural_2018, shain_fmri_2020, ficco_disentangling_2021, schrimpf_neural_2021, clark_surfing_2016, jiang_predictive_2022, wang_predictive_2023, caucheteux_evidence_2023}. It is often seen as part of `active inference', a broader umbrella term which aims to provide an integrated perspective on brain, cognition, and behavior \cite{parr_active_2022}. 

Although PC was partly inspired by ML methods \cite{dayan_helmholtz_1995, rao_correlates_1997, rao_predictive_1999}, within the ML community it appears to have gone mostly unnoticed until recently. In 2017, Whittington \& Bogacz applied PCNs to supervised learning tasks, revealing notable properties that connect them to traditional feedforward neural networks (FNNs). First, PCNs become \textit{equivalent} to FNNs {during inference} (testing) \cite{whittington_approximation_2017, song_predictive_2021, frieder_non-convergence_2022, van_zwol_predictive_2024}. Second, PCN's training algorithm, called inference learning (IL) can be related to backpropagation (BP), the workhorse training algorithm of modern deep learning. These properties sparked a surge of interest in PCNs, suggesting IL as a more biologically plausible alternative for training deep learning architectures \cite{song_can_2020, salvatori_predictive_2021, salvatori_reverse_2022, millidge_predictive_2022, millidge_theoretical_2023}.

Recent work \cite{song_inferring_2024} showed that IL differs from BP by using \textit{prospective configuration}. According to this principle, neural activities across the network change before synaptic weights are modified, in order for neurons to better predict future inputs -- making activity changes `prospective'. \cite{song_inferring_2024} showed that this principle naturally explains several patterns of neural activity and behavior in human and animal learning experiments which are unexplainable by BP --  providing additional evidence for its biological plausibility. This property has been linked to IL's sensitivity to second-order information in the loss landscape \cite{innocenti_understanding_2023, alonso_understanding_2024}, which leads to faster convergence and enhanced performance on certain learning tasks, like continual learning and online learning \cite{song_inferring_2024, alonso_theoretical_2022, ororbia_neural_2022, innocenti_understanding_2023}.

Until recently, implementations of PCNs were more computationally intensive than equivalent BP-trained networks, helping to explain the lack of wider adoption of IL in ML applications. 
However, recent work has achieved marked improvements in performance, showing it can achieve higher efficiency than BP with sufficient parallelization \cite{salvatori_stable_2024, alonso_understanding_2024}. This is because IL's computations use only locally available information, such that the serial updates inherent to BP can be avoided. If sufficiently parallelized, this means that computation time no longer scales with depth in PCNs \cite{salvatori_stable_2024}. As FNNs become increasingly large (deep), this could provide a marked advantage compared to BP, and further suggests PCNs are a promising candidate for use on neuromorphic hardware \cite{<millidge_predictive_towards>, salvatori_stable_2024}.


While recent work has compared PCNs to FNNs, in particular focusing on supervised learning, fundamentally the PCN is a probabilistic (Bayesian) model, naturally formulated for \textit{unsupervised} learning. 
In fact, even before it was used within ML, the PCN of Rao \& Ballard was  phrased in terms familiar to the ML community \cite{rao_predictive_1999, marino_predictive_2022}.  
Specifically, it is defined by a graphical model with latent (unobserved) variables, trained using generalized Expectation Maximization \cite{friston_learning_2003, bishop_pattern_2006}, a general procedure for maximum likelihood in latent variable models (also used, e.g., in $k$-means clustering) \cite{dempster_maximum_1977, neal_view_1998, bishop_pattern_2006}. Its objective (i.e., cost function) is the complete data log-likelihood or variational free energy when seen as a variational inference procedure \cite{friston_theory_2005, friston_free-energy_2010, bishop_pattern_2006}. Hence, from this perspective, PCNs are most appropriately compared not to FNNs, but to techniques associated with \textit{generative modeling} (e.g., variational autoencoders, diffusion models \cite{prince_understanding_2023}) and classic latent variable models (e.g., probabilistic PCA \cite{bishop_pattern_2006}, factor analysis \cite{goodfellow_deep_2016}). 
In other words, PC can be seen \textit{both} as a {learning algorithm} contrasted with BP, and as a {probabilistic latent variable model} comparable to generative models. 



\begin{figure}[]
  \centering
  \includegraphics[width=1.0\textwidth]{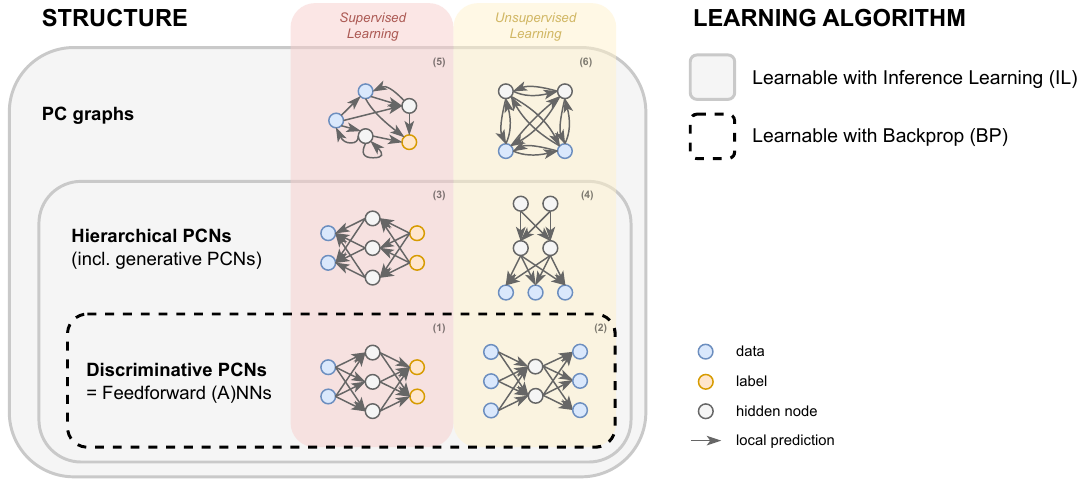}
  \caption{\textbf{Overview of Predictive Coding {N}etworks (PCNs)}. PCNs provide a flexible framework with innovation in {structure} (generative PCNs, PC graphs) and {learning algorithm} (inference learning), compared to traditional ANNs trained with backprop. Examples of network types are schematically illustrated. The structure of PCNs and PC graphs form supersets of (feedforward) ANNs. Schematic architectures shown are: (1) multilayer perceptron/standard FNN architecture, (2) autoencoder (feedforward), (3) supervised generative PCN, (4) unsupervised generative PCN, (5) arbitrary PC graph, (6) fully connected PC graph.}
  \label{fig:overview}
\end{figure}

In PCNs, the difference between supervised and unsupervised learning comes down to the {direction of predictions} within the network. In the supervised setting, these flow from data to the labels, while in the unsupervised setting, they flow towards the data (fig. \ref{fig:PC_original}). Recent work \cite{salvatori_learning_2022} showed how this direction can be extended to a broader notion of \textit{topology} by defining IL on \textit{any graph}, known as PC graphs. This allows the training of non-hierarchical, brain-like structures, going beyond hierarchical to {heterarchical} prediction. This means PCNs effectively \textit{encompass} FNNs, allowing the definition of a more general class of neural networks. {With a hierarchical structure, these networks share with FNNs the appealing property of being universal function approximators \cite{van_zwol_predictive_2024}. } Simultaneously, they allow a new collection of structures untrainable with BP to be studied. Although the study of such structures is in its infancy \cite{salvatori_predictive_2024}, this means they form a {superset} of traditional ANNs (fig. \ref{fig:overview}) {\cite{van_zwol_predictive_2024}.} 

Fig. \ref{fig:equilateral_triangle} shows three different perspectives on PCNs that can be found in the literature. Underlying these is the {mathematical structure} of PCNs, as originally outlined in \cite{friston_hierarchical_2008, friston_free-energy_2010, buckley_free_2017}. Substantial progress has been made since these works, however, both theoretically and in applying PCNs to larger datasets. In addition, no up-to-date pedagogic treatment (tutorial) currently exists. Hence, there is a need for a more comprehensive formal specification of PCNs, and a tutorial aimed at ML practitioners. Our work aims to provide both in the form of a combined tutorial/survey paper, focusing on PC as a {neuro-inspired branch of ML}. This means we choose not to discuss the large body of work on PC in neuroscience, psychology, philosophy, active inference, and robotics (see aforementioned references and \cite{lanillos_active_2021, parr_active_2022} for reviews). Specifically, we do not discuss the neurologically-relevant issue of biological plausibility, often discussed in the PC literature (see e.g. \cite{golkar_constrained_2022, song_inferring_2024}).

\begin{figure}[ ]
\centering

\begin{tikzpicture}[scale=2]
\coordinate (A) at (0,0);
\coordinate (B) at (1,0);
\coordinate (C) at (60:1);

\fill[gray!20] (A) -- (B) -- (C) -- cycle;

\draw (A) -- (B) -- (C) -- cycle;

\node[align=center,below left] at (A) {{\textbf{Probabilistic latent}}\\{\textbf{variable model}}\\\textit{Expectation maximization}\\\textit{Variational free energy}};
\node[align=center,below right] at (B) {{\textbf{Learning algorithm}}\\\textit{Inference learning}\\\textit{Prospective configuration}};
\node[align=center,above] at (C) {{\textbf{Generalized ANN}}\\\textit{Discriminative/generative PCNs,}\\\textit{PC graphs}};


\node at (barycentric cs:A=1,B=1,C=1) {\textbf{PCNs}};
\end{tikzpicture}
\caption{\textbf{Perspectives on PCNs}. {The formal structure of PCNs warrants different complementary perspectives: a learning algorithm (e.g., comparable to backprop), a probabilistic latent variable model (e.g., comparable to variational autoencoders), and a generalization of ANNs to arbitrary graphs. } }
\label{fig:equilateral_triangle}
\end{figure}








\needspace{5\baselineskip}
{\subsection*{Related work \& Contribution}}
{In 2017, predating most work on PC in ML, the only tutorial on PCNs was published~\cite{bogacz_tutorial_2017}. 
In 2022 and 2023, three surveys appeared, related to ours:
\begin{enumerate}
    \item Millidge et al.~\cite{<millidge_predictive_towards>} published a ML-oriented survey, providing a concise introduction to our topic. Although thematically closely related, our work is distinct in its detailed formal analysis, including mathematical details. 
    \item Millidge et al.~\cite{millidge_predictive_2022} provides a comprehensive, gentle overview of PC. It differs from ours by considering a broader scope, including perspectives from neurobiology and psychiatry.
    \item Salvatori et al.~\cite{salvatori_brain-inspired_2023} takes a broad scope, positioning PC within ML. Our review differs by its explicit pedagogic (tutorial) approach, and greater depth regarding canonical PCNs. 
\end{enumerate}}

{
In summary, the contributions of this work are as follows:
\begin{itemize}
    \item Provide a comprehensive formal specification of modern PCNs.
    \item Summarize and provide structure for recent theoretical and empirical results on PCNs, integrating different perspectives (fig.~\ref{fig:equilateral_triangle}), as both a unified basis for future work, and starting-point for ML practitioners. Our aim is to provide a technical reference while remaining accessible.
    \item Explain connections with existing methods in ML, some of which have remained unexposed in recent literature. In particular, we highlight how the {structure} of PCNs and PC graphs forms a mathematical {superset} of ANNs. This follows from earlier work \cite{salvatori_learning_2022}; but, to our knowledge was not yet pointed out as a general conception of PCNs.
\end{itemize}}
{Also, we provide an accompanying Python library (\textbf{\texttt{PRECO}}\footnote{Available at \url{https://github.com/bjornvz/PRECO}.}) using PyTorch that implements PCNs and PC graphs, as a hands-on tutorial.}


\subsection*{Overview}
The three perspectives in fig.~\ref{fig:equilateral_triangle} define the structure of this work. While interconnected, making a distinction facilitates an understanding of PCNs. Section~\ref{sec:ANN_to_PCN} interprets PCNs as {generalized ANNs} (fig.~\ref{fig:overview}), which should provide a familiar context for many ML researchers. Section~\ref{sec:probmodel} discusses PCNs as {probabilistic latent variable models} and Section~\ref{sec:learning} discusses work focusing on IL: PC as a learning algorithm.
The literature on PCNs is organized in Table~\ref{tab:refs} by network type (Section~\ref{sec:ANN_to_PCN}), which we survey throughout this work. We conclude in Section~\ref{sec:conclusion}, with appendices including an extended discussion of Section~\ref{sec:probmodel} in Appendix~\ref{sec:app_Bayes},
complementary proofs in Appendix~\ref{sec:app_proof_FNN_PCN}, and a discussion of computational complexity in Appendix~\ref{sec:app_CC}.

\begingroup
\setcellgapes{4pt} 
\makegapedcells

\begin{table}[]
\centering
\caption{\textbf{Overview of references surveyed in this work}.}
\makegapedcells
\resizebox{\columnwidth}{!}{%
\begin{tabular}{llll}
\hline
\textbf{Subject} &
   &
  \textbf{References} &
  \textbf{Section} \\ \hline
Background &
  Existing reviews &
  \cite{bogacz_tutorial_2017, millidge_predictive_2022, <millidge_predictive_towards>, marino_predictive_2022, salvatori_brain-inspired_2023} &
  \multirow{2}{*}{\ref{sec:intro}} \\
 &
  PhD theses &
  \cite{millidge_applications_2021, song_predictive_2021, <salvatori_thesis>, alonso_energy-based_2023} &
   \\ \hline
Discriminative PCNs &
  IL's relation to BP &
  \cite{whittington_approximation_2017, song_can_2020, salvatori_predictive_2021, millidge_predictive_2022, rosenbaum_relationship_2022, salvatori_reverse_2022, zahid_curvature-sensitive_2023, millidge_backpropagation_2023} &
  \ref{sec:ANN_to_PCN}, \ref{sec:learning} \\
 &
  IL in its natural regime &
  \cite{salvatori_brain-inspired_2023, millidge_theoretical_2023, innocenti_understanding_2023, song_inferring_2024, frieder_non-convergence_2022, millidge_backpropagation_2023, frieder_bad_2024, salvatori_stable_2024, alonso_understanding_2024, mali_tight_2024, pinchetti_benchmarking_2024, ishikawa_local_2024, innocenti_only_2024, innocenti_mupc_2025} &
   \\
 &
  Extensions &
  \cite{salvatori_learning_2022, salvatori_stable_2024, byiringiro_robust_2022, pinchetti_predictive_2022, salvatori_predictive_2024, van_zwol_predictive_2024} &
   \\ \hline
Generative PCNs &
  Supervised learning &
  \cite{kinghorn_preventing_2023, salvatori_learning_2022} &
  \multirow{2}{*}{\ref{sec:generative_PCN}, \ref{sec:probmodel}} \\
 &
  Unsupervised learning &
  \cite{rao_predictive_1999, salvatori_associative_2021, ororbia_neural_2022, ofner_generalized_2022, zahid_predictive_2023, zahid_sample_2024} &
   \\ \hline
Other variations and applications &
  Memory models &
  \cite{tang_recurrent_2023, tang_sequential_2023, li_modeling_2023} &
  \multirow{3}{*}{} \\
 &
  Temporal prediction &
  \cite{jiang_dynamic_2024, millidge_predictive_2024} &
   \\
 &
  Other &
  \cite{boutin_iterative_2020, tscshantz_hybrid_2023, salvatori_predictive_2024} &
   \\ \hline
Other uses of PC in ML &
  Neural generative coding (NGC) &
  \cite{ororbia_neural_2022, ororbia_lifelong_2022, ororbia_backprop-free_2022, ororbia_convolutional_2023, ororbia_spiking_2023} &
  \multirow{2}{*}{} \\
 &
  PC-Inspired ANNs &
  \cite{chalasani_deep_2013, lotter_deep_2017, han_deep_2018, ye_anopcn_2019, rathjens_classification_2024, choksi_brain-inspired_2020, lotter_neural_2020, choksi_predify_2021, pang_predictive_2021, alamia_role_2023, faye_mathematical_2023, chen_cogdpm_2024, rao_active_2023, rao_active_2023-1} &
   \\ \hline
\end{tabular}
}
\label{tab:refs}
\end{table}

\endgroup

\section{PCNs as Generalized ANNs}\label{sec:ANN_to_PCN}
The theory of PC as a machine learning algorithm is often derived from variational inference in a {hierarchical Gaussian latent variable model}. The resulting equations bear a lot of similarity to those of artificial neural networks (ANNs), however. In particular, PCNs of a particular type, sometimes called \textit{discriminative PCNs}, are in an important sense equivalent to feedforward neural network \textit{during inference} (testing). Since we believe ANNs to be more familiar context for most ML researchers and practitioners than variational inference, we will here take ANNs as a starting point, with the aim of making the generalization to PCNs more transparent. 

Thus, our discussion introduces discriminative PCNs by analogy with the simplest ANN: a feedforward neural network (FNN; also known as a multilayer perceptron) in a supervised learning context. Similarities and differences are discussed step by step.\footnote{At this point, we note briefly that a second usage of term `PCN' exists, referring to backprop-trained ANNs with an architecture that \textit{takes inspiration from} PC, e.g. includes separate top-down and bottom-up FNNs (cf. section \ref{sec:PC-ANNs} and `PC-inspired ANNs' in table \ref{tab:refs}).}

\paragraph{Problem setup.} We are given a dataset of $N$ labeled samples $\{\bm{x}^{(n)},\bm{y}^{(n)}\}_{n=1}^N$ split into a training set and a test set, where $\bm{x}^{(n)}\in X$ is a datapoint and $\bm{y}^{(n)} \in Y$ its corresponding label, defined on input and output domains $X$, $Y$, respectively \cite{bishop_pattern_2006}.

\subsection{From FNNs to PCNs}\label{sec:discriminative_PCNs}

\subsubsection{The Activity Rule}
FNNs are defined by a set of \textit{activation nodes} organized in a hierarchy of $L$ layers: $\bm{a}^\ell \in \mathbb{R}^{n_\ell+1}$,  where $\ell=0,...,L$ is the layer and $n_\ell$ the number of nodes in that layer (a constant value of 1 is added to account for the bias). Layers are connected by a set of {weight matrices} $\bm{w}^\ell \in \mathbb{R}^{ n_{\ell+1} \times (n_{\ell}+1)} $ in the following way \cite{bishop_pattern_2006, goodfellow_deep_2016}: 
\begin{equation}
\bm{a}^{\ell}=f(\bm{w}^{\ell-1} \bm{a}^{\ell-1}),
\label{eq:activity_FNN}
\end{equation}
where throughout, the biases $\bm b^{\ell}$ are absorbed as an additional column in the weight matrix (as is commonly done \cite{goodfellow_deep_2016}). We will call this equation the \textit{activity rule} since it states how activation nodes are computed. The function $f: \mathbb{R}^{n_\ell} \rightarrow \mathbb{R}^{n_\ell}$  here is a nonlinear {activation function} (e.g. sigmoid, ReLU). Setting the bottom layer $\bm{a}^0$ to a datapoint $\bm{x}^{(n)}$, the final layer $\bm{a}^L$ is taken as the predicted label $\hat{\bm{y}}=\hat{\bm{y}}(\bm{w},\bm{x}^{(n)})$. 

In PCNs, one similarly has nodes $\bm{a}^\ell$, but the activity rule is quite different. In line with the fundamental notion of PC, each activity in layer $\ell$ is `predicted' by layer $\ell-1$. The \textit{local prediction} is defined as
\begin{equation}
\bm{\mu}^\ell=f(\bm{w}^{\ell-1}\bm{a}^{\ell-1})
\label{eq:prediction}
\end{equation}
(not to be confused with the `global' prediction $\hat{\bm{y}}$ of the network as a whole). {The box {on page~\pageref{sec:direction}} discusses other conventions used in defining the local prediction.}
Here, weights and biases are defined just like in FNNs. The discrepancy between actual and predicted activity is
\begin{equation}
    \bm{\epsilon}^\ell=\bm{a}^{\ell}-\bm{\mu}^\ell,
\label{eq:error}
\end{equation}
i.e. the \textit{prediction error}. One then defines an \textit{energy} function\footnote{In the probabilistic latent variable model perspective (section \ref{sec:probmodel}) this is the \textit{complete data log-likelihood}. Section \ref{sec:ebms} further discusses the use of the term `energy'.} as the sum of squared prediction errors: 
\begin{equation}
    E(\bm{a}, \bm{w}) = \frac{1}{2} \sum_\ell (\bm{\epsilon}^\ell)^2.
\label{eq:energy}
\end{equation}
Training is now done by setting the bottom layer $\bm{a}^{0}$ to the datapoint $\bm{x}^{(n)}$, like in FNNs, but additionally fixing the final layer $\bm{a}^{L}$ to the correct label $\bm{y}^{(n)}$. The activity rule is not defined as in \eqref{eq:activity_FNN}; instead, the updated values of the hidden (unclamped) activity nodes (i.e. $\ell = 1,...,L-1$) are the minimum of the energy function:
\begin{equation}
\hat{\bm{a}}^\ell = \underset{\bm{a}^\ell}{\operatorname{argmin}} \, {E}(\bm{a}, \bm{w}).
\label{eq:activity_PCN}
\end{equation}
Finding this minimum is called the \textit{inference phase}, since we are \textit{inferring} the appropriate activation values for hidden nodes, given the clamped nodes. Typically, finding the minimum $\hat{\bm a}^\ell$ requires gradient descent (an important exception is the testing of discriminative PCNs, as discussed below). Taking the derivative of \eqref{eq:energy}, one obtains:
\begin{equation}
\begin{aligned}
\Delta \bm{a}^\ell&=-\gamma\frac{\partial E}{\partial\bm{a}^\ell}\\
&=-\gamma\left( \bm{\epsilon}^\ell-(\bm{w}^\ell )^T \bm{\epsilon}^{\ell+ 1}\odot f'( \bm{w}^\ell\bm{a}^\ell ) \right).
\end{aligned}\label{eq:a_update}
\end{equation}
Here, $\gamma$ is the \textit{inference rate}: a step size required by gradient descent (i.e., the learning rate for activation nodes), and we use a hat to denote the optimized values. As such, to find $\hat{\bm{a}}^\ell$ during training, one does $T$ iterations of \eqref{eq:a_update} until convergence. This might suggest that training a PCN is more computationally costly than FNNs. However, this added cost can be at least partially circumvented in practice, as we discuss below. 

\subsubsection{The Learning Rule}
In FNNs, the produced output $\hat{\bm{y}}$ is compared to the true label $\bm{y}^{(n)}$ using a {loss function} $\mathcal{L}(w)$ (e.g. the MSE loss).
The optimal weights are those that minimize the loss function, which defines the \textit{learning rule} for ANNs: $\,\hat{\bm{w}}^\ell = \operatorname{argmin}_{\bm{w}} \, \mathcal{L}(\bm{w})$.
This can be found using gradient descent using a learning rate $\alpha$, which can be shown (appendix \ref{sec:app_BP}) to give
\begin{equation}
\begin{aligned}
    \Delta \bm{w}^\ell&=-\alpha\pdv{\mathcal{L}}{\bm{w}^\ell}\\
    &=\alpha \bm{\delta}^{\ell+1}\odot f'(\bm{w}^\ell\bm{a}^\ell)(\bm a^\ell)^T\,,
\label{eq:BP_error1-5}
\end{aligned}
\end{equation}
where $\bm{\delta}^{\ell}$ is the \textit{error} in layer $\ell$, given by 
\begin{equation}
    \bm{\delta}^\ell\equiv\frac{\partial\mathcal{L}}{\partial \bm{a}^\ell} =\begin{dcases}\bm{a}^L-\bm{y}^{(n)}\quad &\ell=L \\
(\bm{w}^\ell)^T\bm{\delta}^{\ell+1}\odot f'(\bm{w}^\ell\bm{a}^\ell)\quad &\ell< L\end{dcases}.
\label{eq:BP_errpr}
\end{equation}
Observe how computing weight updates, then, requires computing the error at the output $\ell = L$, and sequentially computing errors in the other layers in a backwards fashion. This, of course, is what gives the name to \textit{backpropagation} (BP). We show pseudocode for BP in alg. \ref{alg:BP}.
\begin{algorithm}
\caption{Learning $\{\bm x^{(n)}, \bm y^{(n}\}$ with BP}\label{alg:BP}
\begin{algorithmic}[1]
\REQUIRE: $\bm a^0 = \bm x^{(n)}$ .
\FOR {$\ell=0$ to $L-1$} \COMMENT{Forward pass}
\STATE $\bm{a}^{\ell+1}= f(\bm{w}^{\ell}\bm{a}^{\ell})$ 
\ENDFOR
\STATE $\bm{\delta}^L=\pdv{\mathcal{L}}{\bm a^L}$
\FOR {$\ell=L-1$ to $0$} \COMMENT{Backward pass}
\STATE $\bm{\delta}^\ell=(\bm{w}^\ell)^T\bm{\delta}^{\ell+1}\odot f'(\bm{w}^\ell\bm{a}^\ell)$
\STATE $\bm{w}^\ell \gets \bm{w}^\ell-\alpha\pdv{\mathcal{L}}{\bm{w}^\ell}$ \COMMENT{Weight update}
\ENDFOR
\end{algorithmic}
\end{algorithm}

In PCNs, optimal weights are found in a similar fashion. However, one does not minimize a loss function, but once again the energy function -- the same as used for the activity rule -- evaluated at the inference equilibrium: \mbox{$\,\hat{\bm{w}}^\ell = {\operatorname{argmin}}_{\bm{w}^\ell} \, E(\hat{\bm{a}},\bm{w}) $}.
Taking the derivative of \eqref{eq:energy} yields:
\begin{equation}
\begin{aligned}
  \Delta \bm{w}^\ell &=-\alpha\pdv{{E}}{\bm{w}^\ell}\\
&= \alpha \bm{\epsilon}^{l+1}\odot f'(\bm{w}^\ell\hat{\bm{a}}^\ell)(\hat{\bm{a}}^\ell)^T.
  \label{eq:PCN_weight_update}
\end{aligned}
\end{equation}
Note that, although this is update rule is very similar to \eqref{eq:BP_error1-5}, because errors are computed locally \eqref{eq:error}, there is no sequential propagation of errors as in \eqref{eq:BP_errpr}. Thus, once inference has converged, weight updates can be directly computed using inferred activations in neighboring layers.

\begin{mdframed}[backgroundcolor=lightergray]
\subsection*{Conventions}\label{sec:direction}
{Having introduced the defining equations of PCNs, let us mention two issues of convention to assist readers in navigating the PC literature.}

\subsubsection*{Direction}
 The literature contains conflicting notions of \textit{direction} in PCNs, which are worth discussing at the outset. There are two connected issues, the first being mathematical: should local predictions $\bm \mu^{\ell}$ be defined as $\bm \mu^{\ell}=f(\bm w^{\ell+1}\bm a^{\ell+1})$ (implied by fig. \ref{fig:PC_original}) or $\bm \mu^{\ell}=f(\bm w^{\ell-1}\bm a^{\ell-1})$ (as in this section)? The second issue is terminological: how should the words \textit{forward/backward} be used?

Typically in the neuroscientific PC literature \cite{rao_predictive_1999, friston_hierarchical_2008}, predictions are $\bm \mu^{\ell}=f(\bm w^{\ell+1}\bm a^{\ell+1})$, and data is clamped to $\ell=0$ (fig. \ref{fig:PC_original}). This means that predictions go \textit{towards data} (and errors go away from data). In this {`PC convention'}, the word \textit{forward} is typically chosen to mean {away from the data}, which is the direction of errors \cite{rao_predictive_1999}. However, in the supervised learning context, for a PCN to be equivalent to FNNs during testing, it turns out that one needs to have \textit{predictions} go {away from data}. With the conventions above, this means one would have to clamp data to the highest layer $\ell=L$, which is confusing from the perspective of standard ML. This also leads to ambiguity as to whether `forward' should mean towards increasing layer number, or away from the data. 

The alternative when comparing PCNs to FNNs is to keep data clamped to $\ell=0$ but \textit{change the direction of local prediction} to $\bm \mu^{\ell}=f(\bm w^{\ell-1}\bm a^{\ell-1})$, as is done by many recent works \cite{millidge_predictive_2022, rosenbaum_relationship_2022, salvatori_reverse_2022, zahid_predictive_2023, millidge_backpropagation_2023, alonso_theoretical_2022, millidge_theoretical_2023, innocenti_understanding_2023, song_inferring_2024, alonso_understanding_2024, byiringiro_robust_2022, pinchetti_predictive_2022}. \textit{Forward} then means away from the data (equivalently, towards higher layers) as in the PC convention above, except that this is now the direction of predictions. In other words, rather than swapping the data clamping from $\ell=0$ to $L$, we instead swap the directions of predictions and errors relative to fig. \ref{fig:PC_original}, so that -- when comparing PCNs to FNNs specifically -- predictions go forwards and errors go backwards.

The disadvantage of this choice is that $\bm \mu^{\ell}=f(\bm w^{\ell+1}\bm a^{\ell+1})$ with data at $\ell=0$ remains the more sensible convention for PCNs when seen as Bayesian/generative models (section \ref{sec:probmodel}) \cite{rao_predictive_1999, salvatori_associative_2021, ororbia_neural_2022, ofner_generalized_2022}; it is only in the comparison with FNNs where some inversion (either of clamping or of directions) is required. Hence, in this work we use both conventions \textit{depending on the use of the network}. We define \textit{discriminative PCNs} as $\bm \mu^{\ell}=f(\bm w^{\ell-1}\bm a^{\ell-1})$ (for supervised learning), and \textit{generative PCNs} as $\bm \mu^{\ell}=f(\bm w^{\ell+1}\bm a^{\ell+1})$ (for unsupervised learning). As for terminology, we can then unambiguously use `forward' to mean \textit{away from the data} in all cases, consistent with standard ML. One must simply bear in mind that for discriminative PCNs, `forward' reflects the direction of predictions, while for generative PCNs it refers to the direction of errors.
\subsubsection*{Order of Weights and Activation}
{The second convention issue relates to the order in which one applies the weights and non-linear activation function to produce predictions.} We here use $\bm \mu^\ell = f(\bm w^{\ell-1}\bm a^{\ell-1} + \bm b^{\ell-1})$, which is a common convention in ML \cite{goodfellow_deep_2016, prince_understanding_2023} and in line with original work on PC \cite{rao_predictive_1999}. Many works in the PC literature however, define predictions as $\bm \mu^\ell = \bm w^{\ell-1}f(\bm a^{\ell-1})+\bm b^{\ell-1}$ (the \texttt{pypc} library uses yet another convention: $\bm \mu^\ell = f(\bm w^{\ell-1}\bm a^{\ell-1}) + \bm b^{\ell-1}$ \cite{tschantz_infer-activelypypc_2023}). This choice often does not make a difference for performance, but it does change the form of the update rules, both for activations \eqref{eq:activity_PCN}, and weight \eqref{eq:PCN_weight_update}. Moreover, in some cases the choice of convention \textit{is} of great practical importance, since it determines the domain of possible outputs (e.g., a sigmoid or softmax is appropriate for classification but fatal for regression).
\end{mdframed}

The combination of the activity rule \eqref{eq:activity_PCN}, learning rule \eqref{eq:PCN_weight_update}, and energy \eqref{eq:energy}, thus defines a learning algorithm distinct from backprop. It is sometimes called \textit{inference learning} (IL) \cite{salvatori_associative_2021, alonso_theoretical_2022, alonso_understanding_2024}, or, rarely,\textit{ prospective configuration} (cf. section \ref{sec:prosp_config}), but is often simply referred to as `predictive coding' \cite{salvatori_learning_2022, salvatori_stable_2024}. Importantly, it is an instance of \textit{expectation maximization} (EM) (cf. section \ref{sec:probmodel}). We choose the term IL in this work, and show pseudocode in alg. \ref{alg:IL}. 

\begin{algorithm}[h]
\caption{Learning $\{\bm x^{(n)}, \bm y^{(n}\}$ with IL.}\label{alg:IL}
\begin{algorithmic}[1]
\REQUIRE: $\bm a^0 = \bm x^{(n)}$, $\bm a^L = \bm y^{(n)}$  \COMMENT{Clamp data}
\REQUIRE: $\bm a^\ell(0)=\bm \mu^\ell$ for $\ell =1, \ldots, L-1$ \COMMENT{Feedforward initialization}
\FOR {$t=0$ to $T$} \COMMENT{Inference}
\FOR {each $\ell$} 
\STATE $\bm{a}^\ell({t+1})= \bm{a}^\ell(t)-\gamma\pdv{E}{\bm{a}^\ell}$ \COMMENT{Activation update}
\ENDFOR
\ENDFOR
\FOR {each $\ell$} 
\STATE $\bm{w}^\ell \gets \bm{w}^\ell-\alpha\pdv{E}{\bm{w}^\ell}$ \COMMENT{Weight update}
\ENDFOR
\end{algorithmic}
\end{algorithm}


\subsubsection{Testing Procedure}\label{sec:testing_PCNs}
Testing a FNN simply means producing an output $\hat{\bm y}$ for a given input $\bm x^{(n)}$ using \eqref{eq:activity_FNN}. In a PCN, one should \textit{infer} the correct values of the output $\bm a^L$; one should apply the activity rule \eqref{eq:activity_PCN} with the final layer unclamped. Interestingly, this turns out to be much simpler than during training. With the output nodes unclamped, \eqref{eq:activity_PCN} has an analytical solution.
One can show \cite{whittington_approximation_2017, song_predictive_2021} (see appendix \ref{sec:app_proof_FNN_PCN}) that with $\bm{a}^0=\bm{x}^{(n)}$, the minimum of $E$ is: 
\begin{equation}
    \hat{\bm{a}}^1=f(\bm{w}^0 \bm{x}^{(n)})~,\quad
    \hat{\bm{a}}^2=f(\bm{w}^1 \hat{\bm{a}}^1)~,\quad
    \ldots\quad
    \hat{\bm{a}}^L=f(\bm{w}^{L-1} \hat{\bm{a}}^{L-1}).
    \label{eq:discr_PCN_testing}
\end{equation}
{An intuition for this is that during testing, there is one less constraint; {the output is no longer connected to the data, so all the network can do is use the learned weights to produce an output.}} In other words, testing in a PCN equates to a single pass through the network, not requiring gradient-based optimization of activation nodes \eqref{eq:a_update}. In other words, we can say that testing is equivalent in PCNs and FNNs, or that during testing the PCN `becomes' an FNN in the sense of \eqref{eq:activity_FNN} \cite{van_zwol_predictive_2024}. As such, important results for neural networks such as universal approximation theorems \cite{prince_understanding_2023} also hold for discriminative PCNs. Indeed, we emphasize that this holds only for \textit{discriminative} PCNs, with the local prediction defined as in \eqref{eq:prediction}, and data clamped to $\bm{a}^0$. If this is changed, as in later sections, then this equivalence no longer holds.

\subsection{Extensions}

\subsubsection{Other Layers \& Objective Functions}
It is straightforward to extend discriminative PCNs to more complex architectures. E.g., importing convolutional layers to PCNs only requires changing pre-activations (the argument of the nonlinear function) of local predictions, i.e.
\begin{equation}
\mu^{\ell}_{ij}=f\Big( \sum_{x=1}^{k_\ell} \sum_{y=1}^{k_\ell} W_{xy}^{\ell} a_{i+x,\,j+y}^{\ell-1}+b^\ell \Big),
\end{equation}
where $W^{\ell}\in\mathbb{R}^{k_\ell\times k_\ell}$ is the {kernel} with $k_\ell$ the {kernel size}, and $x$ and $y$ define the local receptive field (with stride $s_\ell$ and padding $p_\ell$ one has $x\rightarrow s_\ell x-p_\ell$ and $y\rightarrow s_\ell y-p_\ell$). One can introduce pooling layers and recurrent layers by similarly changing only the pre-activation \cite{salvatori_predictive_2021}.

Other energy functions may also be considered \cite{pinchetti_predictive_2022, millidge_theoretical_2023}. The energy $E=\sum_\ell (\bm \epsilon^\ell)^2$, corresponding to the MSE loss in FNNs, can be generalized to any layer-dependent energy $E=\sum_\ell  E_\ell$, where $E_\ell$ may be another function of activations and weights ${E}_\ell= g(a^\ell,a^{\ell-1}) $ \cite{millidge_theoretical_2023}. For instance, instead of an MSE-like error, \cite{pinchetti_predictive_2022} derives a {cross-entropy} error:
$$
E_\ell= \sum_{i=1}^K a^\ell_i\log {\hat{a}^\ell_i}.
$$
Here, $\hat{a}^\ell_i = \operatorname{softmax}[f(\bm w^{\ell-1}\bm a^{\ell-1})]_i$, appropriate for categorical output variables.\footnote{In the probabilistic perspective, this change corresponds to changing from a Gaussian  to a generalized Bernoulli (categorical) distribution.} This softmax function is the ingredient missing from standard PC to implement transformers, implemented in \cite{pinchetti_predictive_2022}. Using a similar framework, the same authors also train variational autoencoders (VAEs) \cite{kingma_auto-encoding_2014} with IL.

\subsubsection{Incremental Inference Learning}
{It turns out that in the inference phase, \eqref{eq:a_update} does not need to converge in order to obtain useful weight updates. Recently, \cite{salvatori_stable_2024} modified IL to update weights after each inference iteration. This is justified by the probabilistic perspective of PCNs, where IL is simply a form of EM. In \cite{neal_view_1998}, it was shown that EM admits the use of partial, incremental steps.} Thus, a minor change in alg. \ref{alg:IL} gives alg. \ref{alg:iPC}, which we refer to as \textit{incremental IL} in this work. {Further context on incremental EM is provided in section \ref{sec:probmodel} and appendix \ref{sec:app_Bayes}.}


\begin{algorithm}
\caption{Learning $\{\bm x^{(n)}, \bm y^{(n}\}$ with Incremental IL.}\label{alg:iPC}
\begin{algorithmic}[1]
\REQUIRE: $\bm a^0 = \bm x^{(n)}$, $\bm a^L = \bm y^{(n)}$  \COMMENT{Clamp data}
\REQUIRE: $\bm a^\ell(0)=\bm \mu^\ell$ for $\ell =1, \ldots, L-1$ \COMMENT{Feedforward initialization}
\FOR {$t=0$ to $T$}
\FOR {each $\ell$}
\STATE $\bm{a}^\ell({t+1})= \bm{a}^\ell(t)-\gamma\pdv{E}{\bm{a}^\ell}$ \COMMENT{Activation update}
\STATE $\bm{w}^\ell \gets \bm{w}^\ell-\alpha\pdv{E}{\bm{w}^\ell}$ \COMMENT{Weight update}
\ENDFOR
\ENDFOR
\end{algorithmic}
\label{ref:alg_iPC}
\end{algorithm}

\subsection{Comparison}\label{sec:comparisons_maintext}
\subsubsection{Structure} 
\begin{figure}[]
  \centering
  \includegraphics[width=\textwidth]{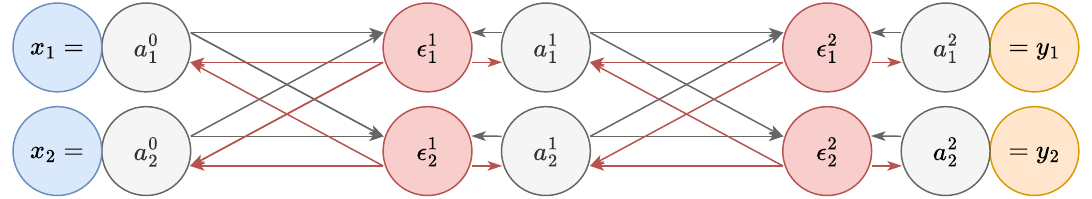}
    \caption{\textbf{Schematic discriminative PCN structure}. A single hidden layer is shown. {Forward connections (gray) correspond to \textit{predictions}, going away from the data, defined by \eqref{eq:prediction} and \eqref{eq:error} (cf. section \ref{sec:direction}). Backward  connections (red) are the direction of \textit{errors}, going towards data, defined by \eqref{eq:a_update}. }The input layer $\bm{a}^{0}$ is clamped to a datapoint $\bm{x}^{(n)}$, and the final layer $\bm{a}^{2}$ to the label $\bm{y}^{(n)}$.}
  \label{fig:PCN}
\end{figure}

A PCN is typically visualized as in fig. \ref{fig:PCN}, which reveals some notable differences with FNNs. PCNs have bottom-up (forward) connections from data to labels (predictions), as well as top-down (backward) connections from labels to data (errors), meaning they have \textit{feedback}/\textit{recurrence}\footnote{It is worth emphasizing that this recurrence is of an altogether different nature than that in recurrent neural networks (RNNs) \cite{goodfellow_deep_2016}, where the recurrence is in time rather than in spatial connectivity (i.e., RNNs are characterized by weight sharing). In PCNs, time steps are iterations in the inference phase, whereas in RNNs, time steps are subsequent samples of temporal data. 
A variation of PCNs applied to temporal data is \cite{millidge_predictive_2024}.} in all nodes, a fundamental feature of PC. 

\subsubsection{Locality \& Computational Complexity}
Importantly, both the activity rule \eqref{eq:a_update} and learning rule \eqref{eq:PCN_weight_update} in IL are {local}: {at each iteration, updating an activity $\bm a^\ell$ or weight $\bm w^\ell$ depends only on quantities in layer $\ell-1,\,\ell,\,\ell+1$. In contrast, in BP an update of $\bm w^\ell$ explicitly depends on activations in layer $\ell,\,\ell+1,\,\dots,\,L$.} 

\begin{figure}[]
  \centering
    \includegraphics[width=0.9\textwidth]{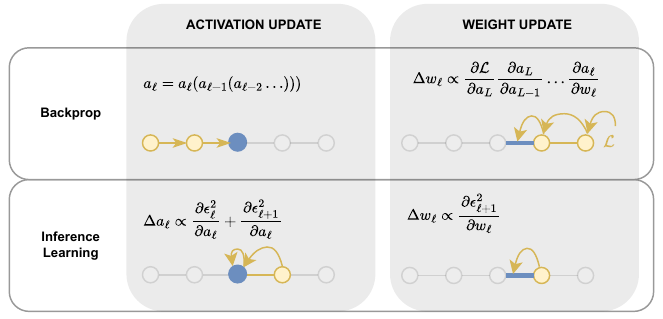}
  \caption{\textbf{Locality in IL.} Schematic illustration of activation and weight updates BP in an FNN vs. IL in a PCN. For the former, each update of activations/weights requires propagating information from all preceding/succeeding activations, respectively. By contrast, in IL each update is computed locally, only requiring information from neighboring layers. This is often cited as a reason for the greater biological plausibility of PCNs, but is practically relevant insofar as it allows the updates to be performed in parallel, and avoids issues surrounding vanishing/exploding gradients. {In the top left, we have left out weights and nonlinearities for simplicity.}}
  \label{fig:locality}
\end{figure}

This means that in IL, updates in all layers can in principle be parallelized -- whereas in BP, updates in lower layers cannot be done before updates in all higher layers have been completed. In other words, BP involves waiting times; it is not local in time. This difference is visualized in fig. \ref{fig:locality}. 


If this parallelization of computations in layers can be realized, this difference implies a potential speed-up of IL compared to BP. This can be analyzed by considering the time complexity of the computations in table \ref{tab:CC_short}, corresponding to the time complexity of \textit{a single weight update}. (Note that reduced time complexity here will not automatically imply lower training time in practice, since the number of weight updates until convergence (typically measured by epochs) depends on a complex interaction of optimizer, dataset, and hyperparameters.)

With matrix multiplications being the most costly computations and defining $M$ as the complexity for the largest weight multiplication, the complexity per weight update without parallelization can be shown to be (cf. appendix \ref{sec:app_CC}) $\mathcal{O}(LM)$ for BP, $\mathcal{O}(TLM)$ for IL, and $\mathcal{O}(LM)$ for incremental IL.\footnote{{Although incremental IL, like standard IL, does $T$ inference iterations (cf. algorithm \ref{alg:iPC}), it also does a weight update at each iteration. Hence, the complexity \textit{per weight update} is only $\mathcal{O}(LM)$ for incremental IL \cite{salvatori_stable_2024, alonso_understanding_2024}, while complexity \textit{per batch} remains $\mathcal{O}(TLM)$. 
}} With parallelization of layers and ignoring potential overhead, IL's time complexity {per weight update} decreases to $\mathcal{O}(TM)$, incremental IL's to  $\mathcal{O}(M)$ (whereas BP's complexity remains unchanged). {Thus, training time no longer explicitly scales with the depth of the network, which is a highly desirable feature (cf. table \ref{tab:CC_short}).}




\begin{table}[]
\caption{\textbf{Time complexity {per weight update} for BP, IL, and incremental IL/incremental PC}. $M$ is the most costly matrix computation, $T$ is the number of steps to convergence, and $L$ is the number of layers (i.e., depth). As discussed above, the local nature of IL allows parallelization to remove the dependence on depth, which is not possible in BP.\label{tab:CC_short}}
\centering
\setcellgapes{3pt}
\makegapedcells
\begin{tabular}{lccc}
\hline
 & \textbf{BP} & \textbf{IL} & \textbf{Incremental IL} \\ \hline
\textbf{Standard} & $\mathcal{O}(LM)$ & $\mathcal{O}(TLM)$ & $\mathcal{O}(LM)$ \\
\textbf{Parallelized} & $\mathcal{O}(LM)$ & $\mathcal{O}(TM)$ & $\mathcal{O}(M)$ \\ \hline
\end{tabular}
\vspace{0.1cm}
\end{table}
Whether this means IL can be faster than BP in practice remains unclear: although a proof of principle exists \cite{salvatori_stable_2024}, its implementation includes substantial computational overhead, and how $T$ scales with network size remains unclear. Moreover, although incremental IL is appealing for its time complexity, it remains unknown whether desirable properties of IL (discussed in \ref{sec:empirical1}, \ref{sec:learning}) transfer to incremental IL. On the other hand, BP is heavily optimized by dedicated libraries that employ GPU acceleration, whereas no similarly comprehensive library yet exists for IL, so larger performance gains are plausible. 






\subsubsection{Objectives}
{The energy used in a PCNs is closely related to the MSE loss in an FNN. With the final layer of the PCN $\bm{a}^L$ fixed to the label $\bm{y}^{(n)}$, writing out $E$ gives:
\begin{equation}
    E(\bm a,\bm w) = \frac{1}{2}(\bm y^{(n)}-\hat{\bm y})^2+\frac{1}{2}\sum_{\ell=0}^{L-1} (\bm \epsilon^\ell)^2,
\end{equation}
where we have labeled the prediction of the last layer $\hat{\bm{y}} = \bm \mu^L\equiv f(\bm{w}^{L-1}\bm{a}^{L-1}) $. Here, the term $\sum_{\ell=1}^{L-1}(\bm \epsilon^\ell)^2$ is sometimes called the \textit{internal error} \cite{millidge_backpropagation_2023} or \textit{residual error} \cite{millidge_theoretical_2023}. Observe that with this notation, the first term looks like the MSE loss for a single datapoint. However (as pointed out in \cite{innocenti_only_2024}) most of the time they are not equal, because $\bm a^{\ell}$ are computed through inference, i.e. \eqref{eq:activity_PCN}, as opposed to the feedforward pass \eqref{eq:activity_FNN}. There are two notable exceptions to this, when the energy \textit{is} precisely equal to the MSE loss. The first is during \textit{testing}, where \eqref{eq:discr_PCN_testing} is equivalent to having zero internal error. The second is directly after a `feedforward initialization', discussed in the next section.}
\subsubsection{Initialization}\label{sec:initialization}
Unlike BP, IL requires \textit{initializing hidden nodes} before inference. Different schemes exist \cite{frieder_non-convergence_2022, ishikawa_local_2024}, such as taking random samples from a probability distribution. Most popular, however, is `feedforward initialization', which means setting $\bm a^\ell(0) = \bm \mu^\ell$ \cite{pinchetti_benchmarking_2024}. This is effectively the feedforward pass of an FNN. This initialization causes the internal error to be zero, with all error concentrated in the final layer---meaning the energy is equal to the MSE loss. During inference this error will be redistributed throughout the network. Although this scheme appears to provide better classification performance than using random initialization \cite{pinchetti_benchmarking_2024} {and is natural to consider for discriminative tasks}, it lacks a clear theoretical basis, and more optimal schemes may exist.


A final note on learning concerns weight initialization. It is well-known in machine learning that in deep networks, correctly initializing weights is required for networks to be trainable \cite{goodfellow_deep_2016, prince_understanding_2023,schoenholz_deep_2017,bukva_criticality_2023}. Some works have started considering this issue in relation to PC \cite{frieder_non-convergence_2022, innocenti_only_2024, ishikawa_local_2024}. {Importantly, recent work \cite{innocenti_mupc_2025} has shown that `parameterizing' weights using Depth-$\mu$P \cite{yang_tensor_2023, bordelon_depthwise_2023} greatly improves performance on ResNets, preventing exploding and vanishing gradients (section \ref{sec:empirical1}).}

\subsubsection{Practical Considerations}
{Until 2024, the only publicly available implementation of PCNs was \texttt{pypc} \cite{tschantz_infer-activelypypc_2023}, but this recently changed with two new libraries based on JAX} \cite{bradbury_jax_2018}: \texttt{PCX} \cite{pinchetti_benchmarking_2024} and \texttt{JPC} \cite{innocenti_jpc_2024}. We accompany this work by \texttt{PRECO}, the only library that also implements PC graphs (cf. Section~\ref{sec:PCgraphs}).

{We briefly comment on the choice of activation function, an important practical consideration when experimenting with PCNs. Namely, recent work has shown that ReLU is a suboptimal choice of activation function for PC~\cite{frieder_bad_2024}, since it creates pathologies in the energy function that prevent the model from learning optimally. Alternatives that typically perform well include tanh, HardTanh and Leaky ReLU~\cite{pinchetti_benchmarking_2024}.} 

\subsection{Empirical Results}\label{sec:empirical1}
As is well known to deep learning practitioners, many factors affect performance of FNNs, which is no different with discriminative PCNs. Exponential growth of the search space with hyperparameters makes a systematic comparison intractable, meaning existing works compare FNNs and PCNs over modest ranges of hyperparameters. Furthermore, although PCNs share many theoretical properties with FNNs, \cite{frieder_non-convergence_2022, frieder_bad_2024} also make clear that care needs to be taken in importing knowledge and intuitions from BP-trained networks since IL does introduce meaningful changes related to its mathematical properties. With this remark we summarize the advantages and limitations that have been observed so far. 

\subsubsection{Small-sized Experiments}
{Most empirical work has been done on relatively small datasets (MNIST \cite{lecun_gradient-based_1998}, F-MNIST \cite{xiao_fashion-mnist_2017}, CIFAR-10 \cite{krizhevsky_learning_2009}) with relatively small models (MLPs or small CNNs) \cite{whittington_approximation_2017, rosenbaum_relationship_2022, alonso_theoretical_2022, pinchetti_predictive_2022, salvatori_stable_2024, song_inferring_2024, pinchetti_benchmarking_2024}. For these cases, the picture that emerges is that IL performs very similarly to BP in terms of accuracy, with sub-percentage differences. Modest gains of single-digit percentages are observed in \cite{alonso_theoretical_2022} and \cite{song_inferring_2024} for specific tasks: {online learning} (learning with batch size 1) and {data efficiency} (learning with less than 300 data points per class), {continual learning} tasks and {concept drift}, with gains of up to 20\% \cite{song_inferring_2024}. However, more comprehensive studies remain to be done to show these findings hold for larger datasets and models.} It is also observed across datasets that IL converges more quickly in terms of epochs \cite{alonso_theoretical_2022, innocenti_understanding_2023,song_inferring_2024, innocenti_only_2024} (cf. section \ref{sec:learning} for theoretical discussion). 

\subsubsection{Larger Experiments}
{{Three} works \cite{salvatori_stable_2024, pinchetti_benchmarking_2024, innocenti_mupc_2025} have performed experiments with larger models and datasets, with \cite{pinchetti_benchmarking_2024} presenting the most comprehensive benchmarking of PCNs in the literature so far.} \cite{salvatori_stable_2024} considers the SVHN dataset \cite{netzer_reading_2011} and networks with up to 8 layers, including an AlexNet architecture \cite{krizhevsky_imagenet_2012}. For the largest architecture, authors observe a modest decrease when using IL, but differences with BP remained small. \cite{pinchetti_benchmarking_2024} considers, next to smaller datasets/models, the CIFAR-100 \cite{krizhevsky_learning_2009} and Tiny ImageNet \cite{krizhevsky_learning_2009} datasets, using VGG-based models \cite{simonyan_very_2015} with 5,7, and 9 layers, and a ResNet model \cite{he_deep_2016} with 18 layers (next to IL and incremental IL, the authors also study number of less studied variants). An interesting picture emerges: IL performs comparably to BP when small {models} are used, but strongly deteriorates when model size increases (cf. table \ref{tab:empirical}). {However, this apparent challenge in scaling IL appears to have been resolved in the recent work by \cite{innocenti_mupc_2025}. The authors argue that what earlier work failed to address was the \textit{instability of the forward pass}, leading to vanishing or exploding gradients in deep networks. By using a scheme called Depth-$\mu$P \cite{yang_tensor_2023, bordelon_depthwise_2023}, the authors show that very deep (100+ layers) ResNets can reliably be trained with IL to competitive performance. More work is needed to establish whether other schemes, such as forms of activity normalization, have the same beneficial effect for IL.}
We remark that faster convergence in terms of epochs, observed as an advantage for smaller datasets, has not yet been studied for larger experiments.



\begin{table}[]
\caption{{\textbf{Test accuracy comparisons between BP and IL for models of increasing depth} on classification tasks. This is a simplified version of table 1 in \cite{pinchetti_benchmarking_2024}, which includes many more variants of IL and databases. To keep this work self-contained, we limit the table the two IL variants discussed in this work; and we have left out some of their results for readability (see \cite{pinchetti_benchmarking_2024} for details on the experiments). {Importantly, contrasting the picture suggested by this table, we note that recently appeared work \cite{innocenti_mupc_2025} achieves performance with IL that equals BP for ResNets {up to 128 layers} by using Depth-$\mu$P.}}   }

\begin{tabular}{|cccc|}
\hline
\multicolumn{1}{|c|}{}           & \multicolumn{1}{c|}{\textbf{IL}}                  & \multicolumn{1}{c|}{\textbf{Incremental IL}}                    & \textbf{BP}                              \\ \hline
\multicolumn{1}{|l}{\textbf{MLP}}                    &                                             &                                             &                                     \\ \hline
\multicolumn{1}{|c|}{MNIST}                 & \multicolumn{1}{c|}{$98.26^{ \pm 0.04}$}    & \multicolumn{1}{c|}{$98.45{ }^{ \pm 0.09}$} & $98.29^{ \pm 0.08}$                 \\ \hline
\multicolumn{1}{|c|}{FashionMNIST}          & \multicolumn{1}{c|}{$89.58^{ \pm 0.13}$}    & \multicolumn{1}{c|}{$89.90^{ \pm 0.06}$}    & $89.48^{ \pm 0.07}$                 \\ \hline
\multicolumn{1}{|l}{\textbf{VGG-5}}                  &                                             &                                             &                                     \\ \hline
\multicolumn{1}{|c|}{CIFAR-10}              & \multicolumn{1}{c|}{$87.988^{ \pm 0.11}$}   & \multicolumn{1}{c|}{$85.51^{ \pm 0.12}$}    & $89.43^{ \pm 0.12}$                 \\ \hline
\multicolumn{1}{|c|}{CIFAR-100 (Top-1)}     & \multicolumn{1}{c|}{$54.08^{ \pm 1.66}$}    & \multicolumn{1}{c|}{$56.07^{ \pm 0.16}$}    & $66.28^{ \pm 0.23}$                 \\ \hline
\multicolumn{1}{|c|}{Tiny ImageNet (Top-5)} & \multicolumn{1}{c|}{$57.31^{ \pm 0.21}$}    & \multicolumn{1}{c|}{$54.73^{ \pm 0.52}$}    & $65.26^{ \pm 0.37}$                 \\ \hline
\multicolumn{1}{|l}{\textbf{VGG-7}}                  &                                             &                                             &                                     \\ \hline
\multicolumn{1}{|c|}{CIFAR-10}              & \multicolumn{1}{c|}{$81.91{ }^{ \pm 0.3}$}  & \multicolumn{1}{c|}{$80.15^{ \pm 0.18}$}    & $89.91{ }^{ \pm 0.1}$               \\ \hline
\multicolumn{1}{|c|}{CIFAR-100 (Top-1)}     & \multicolumn{1}{c|}{$37.52^{ \pm 2.60}$}    & \multicolumn{1}{c|}{$43.99^{ \pm 0.30}$}    & $65.36{ }^{ \pm 0.15}$              \\ \hline
\multicolumn{1}{|c|}{Tiny ImageNet (Top-5)} & \multicolumn{1}{c|}{$44.92{ }^{ \pm 0.27}$} & \multicolumn{1}{c|}{$40.36^{ \pm 0.22}$}    & $66.65^{ \pm 0.20}$                 \\ \hline
\multicolumn{1}{|l}{\textbf{VGG-9}}                  &                                             &                                             &                                     \\ \hline
\multicolumn{1}{|c|}{CIFAR-10}              & \multicolumn{1}{c|}{$75.33^{ \pm 0.25}$}    & \multicolumn{1}{c|}{$79.02^{ \pm 0.21}$}    & 90.02 ${ }^{ \pm 0.18}$             \\ \hline
\multicolumn{1}{|c|}{CIFAR-100 (Top-1)}     & \multicolumn{1}{c|}{$39.57^{ \pm 0.18}$}    & \multicolumn{1}{c|}{$44.76{ }^{ \pm 0.40}$} & $65.51{ }^{ \pm 0.23}$              \\ \hline
\multicolumn{1}{|c|}{Tiny ImageNet (Top-5)} & \multicolumn{1}{c|}{$44.433^{ \pm 0.09}$}   & \multicolumn{1}{c|}{$50.48^{ \pm 0.05}$}    & $65.62^{ \pm 0.17}$                 \\ \hline
\multicolumn{1}{|l}{\textbf{ResNet-18}}              &                                             &                                             &                                     \\ \hline
\multicolumn{1}{|c|}{CIFAR-10}              & \multicolumn{1}{c|}{$53.744^{ \pm 0.43}$}   & \multicolumn{1}{c|}{$70.44^{ \pm 0.81}$}    & ${9 3 . 2 1}{ }^{ \pm 0.07}$ \\ \hline
\multicolumn{1}{|c|}{CIFAR-100 (Top-1)}     & \multicolumn{1}{c|}{$22.83{ }^{ \pm 0.38}$} & \multicolumn{1}{c|}{$29.45{ }^{ \pm 1.36}$} & $71.89{ }^{ \pm 0.16}$              \\ \hline
\multicolumn{1}{|c|}{Tiny ImageNet (Top-5)} & \multicolumn{1}{c|}{$34.55^{ \pm 0.20}$}    & \multicolumn{1}{c|}{$16.51^{ \pm 3.09}$}    & $74.98{ }^{ \pm 0.36}$              \\ \hline
\end{tabular}
\label{tab:empirical}
\end{table}

 \subsubsection{PC-inspired ANNs}\label{sec:PC-ANNs}
{Complementary to PCNs as presented so far, a different line of work exists that uses BP-trained ANNs with architectures that \textit{take inspiration from} PC to improve performance on various tasks (cf. `PC-inspired ANNs' in table \ref{tab:refs}). This alternate use of the term `PCN' includes, e.g., PredNet \cite{lotter_deep_2017, lotter_neural_2020}, contrastive predictive coding \cite{oord_representation_2019}, CogDPM \cite{chen_cogdpm_2024}, and active predictive coding networks \cite{rao_active_2023, rao_active_2023-1}. Although the focus of our work is on models that are not trained with BP, we mention briefly that these models have shown strong performance on several fronts. For instance, \cite{chen_cogdpm_2024} successfully used PC's \textit{precision weighting} (cf. section \ref{sec:precision}) in a diffusion model to substantially improve spatiotemporal forecasting predictions. For further discussion on these models, cf. \cite{millidge_predictive_2022, salvatori_brain-inspired_2023}.}


\subsection{Generative PCNs}\label{sec:generative_PCN}

In the supervised learning context, discriminative models approximate the posterior probability distribution of labels given datapoints, $p(y|x)$ \cite{bishop_pattern_2006}. If equipped with a softmax function at the final layer, this is precisely what the model above does (hence the term discriminative PCN). Making only a minor change to the approach discussed above, one instead obtains a model that approximates $p(x,y)$: a \textit{generative model}, from which synthetic points in the input space can be generated. Generative modeling is typically done using unsupervised learning, concerned with finding patterns and structure in unlabeled data \cite{bishop_pattern_2006}. This means a change in the problem setup: one no longer has a dataset $\{\bm{x}^{(n)},\bm{y}^{(n)}\}_{n=1}^N$, but only $\{\bm{x}^{(n)}\}_{n=1}^N$. {Although generative models have many benefits compared to discriminative models, training them is often more difficult \cite{bishop_pattern_2006, murphy_probabilistic_2023}, and hence their use is less widespread than discriminative models \cite{prince_understanding_2023}.}

To obtain a generative PCN, one changes the structure of the network; specifically, the direction of the local predictions: $\bm{\mu}^\ell=f(\bm{w}^{\ell-1}\bm{a}^{\ell-1})$ becomes $\bm{\mu}^\ell=f(\bm{w}^{\ell+1}\bm{a}^{\ell+1})$.\footnote{Equivalently, one leaves the local predictions unchanged, and swap sides where the data and label are clamped: i.e. clamp $\bm{x}^{(n)}$ to $\bm{a}^{L}$ and $\bm{y}^{(n)}$ to $\bm{a}^{0}$, cf. section \ref{sec:direction}.}
With IL, such a model can be used both for supervised learning, as well as unsupervised learning, with the difference being the training and testing procedures, discussed below.

In general, it should be noted that testing a generative model is less straightforward than testing discriminative models, since they can be used for several purposes (e.g. density estimation, sampling/generation, latent representation learning) \cite{prince_understanding_2023}. 

\subsubsection{Supervised Learning}
Compared to discriminative PCNs, the \textit{training procedure} and \textit{learning algorithm} (IL), are unchanged: \eqref{eq:activity_PCN} is run until converged, followed by a weight update. The changed local prediction results in slight changes to update rules: in \eqref{eq:activity_PCN} and \eqref{eq:PCN_weight_update}, $\ell+1$ becomes $\ell-1$. For the \textit{testing procedure}, clamping the final layer $\bm{a}^{L}$ to a label, with the reversed prediction direction one has the reverse of \eqref{eq:discr_PCN_testing}: a synthetic datapoint is created at $\bm{a}^{0}$ in a single `backward' pass\footnote{In a generative PCN, backward is the direction of predictions (cf. section \ref{sec:direction}), i.e. this backward pass involves the same computations as the forward pass in an FNN.} (without requiring several iterations of \eqref{eq:activity_PCN}). 

\subsubsection{Unsupervised Learning}
 For traditional ANNs, an FNN may be adapted for unsupervised learning by using a special autoencoder architecture, encoding input data into a lower-dimensional representation followed by a decoder which reconstructs the original input. PCNs do not require a special architecture of this sort, but only change the \textit{direction} of prediction to $\bm{\mu}^\ell=f(\bm{w}^{\ell+1}\bm{a}^{\ell+1})$ i.e. the use of a generative PCN. Indeed, PCNs were originally conceived in this way \cite{rao_predictive_1999} (also see section \ref{sec:probmodel}). {This change is straightforward for fully-connected architectures but may require more consideration for, e.g., convolutional layers.} The \textit{training procedure} for unsupervised learning requires only a minor change to the approach described above. Simply keep the final layer $\bm{a}^{L}$ \textit{unclamped} during training, keeping data $\bm{x}^{(n)}$ clamped to $\bm{a}^{0}$, and run IL. The model functions like an \textit{autoencoder}: higher layers take the role of the latent space, which during training encodes increasingly abstract features of the data clamped to the lowest layer. As such, the PCN takes the role of \textit{both} encoder and decoder. {Regularization on the top layer is sometimes added, corresponding to a prior distribution (cf. section \ref{sec:probmodel}) \cite{rao_predictive_1999}.}

As for the \textit{`testing' procedure}, one can sample the latent space as a decoder using a noise vector at the top, {i.e., a prior distribution}, together with \textit{ancestral sampling}. Seeing the PCNs as a hierarchical probabilistic model (cf. section \ref{sec:probmodel}), each layer has a (Gaussian) conditional probability distribution. The idea of ancestral sampling is to generate a sample from the root variable(s) (the top layer) and then sample from the subsequent conditional distributions based on this \cite{prince_understanding_2023}. Finally, a synthetic datapoint at the bottom is obtained \cite{ororbia_neural_2022}. 

\subsubsection{Hierarchical PCNs}
\begin{figure}[h!]
  \centering
  \includegraphics[width=0.95\textwidth]{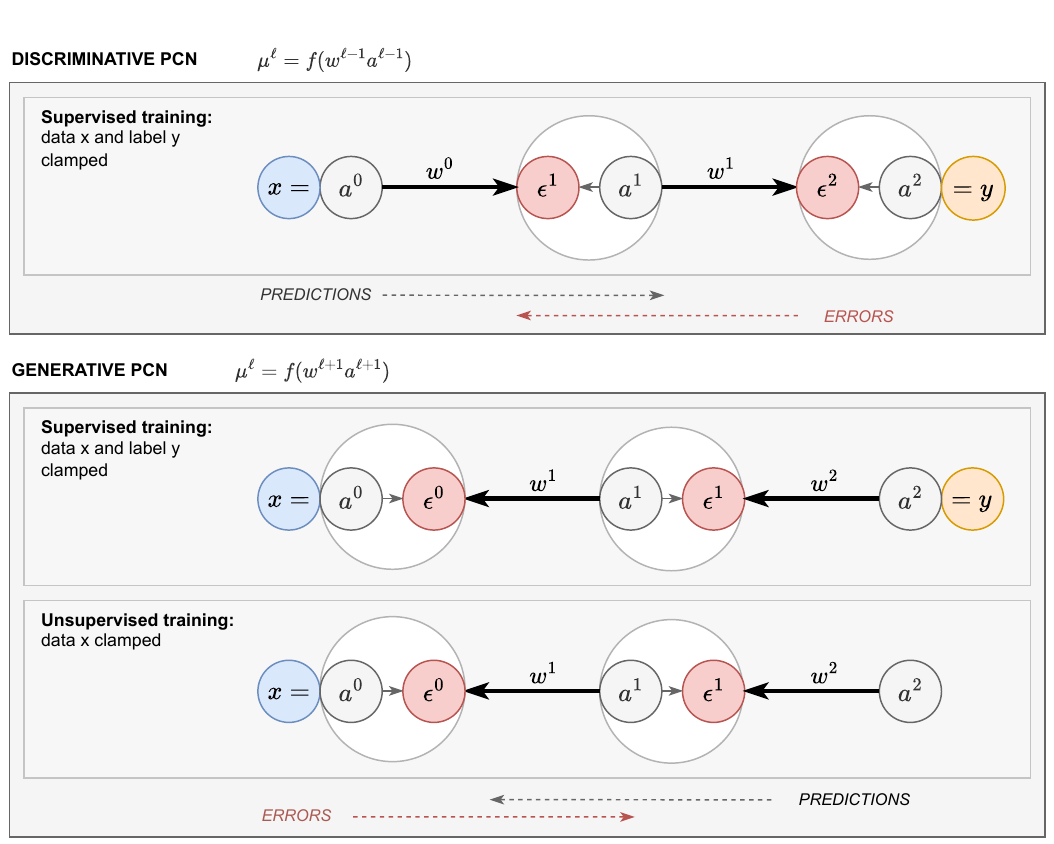}
  \caption{\textbf{Overview of hierarchical PCNs}. Model type (discriminative vs. generative), learning mode (supervised vs. unsupervised), and direction of predictions and errors are shown. Cf. section \ref{sec:direction} for an explanation of our convention. For clarity, one neuron per layer is shown, and bias has been left out. }
  \label{fig:gen_discr}
\end{figure}

Discriminative and generative PCNs can together be called \textit{hierarchical PCNs}, defined by the local prediction $\bm \mu^\ell = f (\bm w^{\ell\pm 1}\bm a^{\ell\pm 1})$. The supervised and unsupervised learning that can be done with generative PCNs can then be called the \textit{training modes} of PCNs. These are illustrated in fig. \ref{fig:gen_discr}. Intuitively, we can see the difference between discriminative and generative PCNs in the direction of predictions: in the discriminative model, predictions flow from data to labels, while errors flow from labels to data. In generative models, this is reversed.\footnote{It should be noted that testing can also be `reversed' in hierarchical PCNs, which we call \textit{backwards testing}. If one has trained the discriminative PCN, one can clamp \textit{labels} and find the minimum energy with iterative inference, producing a synthetic image at the bottom. Conversely, a trained generative PCN can classify images by iterative inference, using a clamped image instead of a clamped label. Thus, confusingly, discriminative PCNs can be used for generative tasks, and generative PCNs can be used for discriminative tasks. However, in practice, neither works better than their forward-tested counterpart \cite{kinghorn_preventing_2023, millidge_predictive_2022}, meaning generative PCN are the natural choice for generation, as discriminative PCNs are for classification. This justifies the naming convention used here and elsewhere in the literature.}

\subsubsection{Empirical Results}
Compared to discriminative PCNs, generative PCNs remain underexplored in the literature, which is perhaps surprising considering the original conception of PC as an unsupervised generative model \cite{rao_predictive_1999}. At the same time, as mentioned, metrics for what counts as a `good' generative model are diverse \cite{prince_understanding_2023}, making comparisons more involved. 

{For completeness, we first mention \cite{salvatori_associative_2021}, which used a generative \textit{model}, but technically not a generative \textit{task}, considering PCNs as associative memories (which does not involve generation of unseen data).} On the task of reconstructing from corrupted and partial data (using up to 500 datapoints), the authors find good performance, outperforming autoencoders, and both Hopfield networks \cite{hopfield_neural_1982} and modern Hopfield networks \cite{krotov_dense_2016} in most cases. 

Here, we review the somewhat disparate results in the literature for generation tasks. {Most empirical  works have compared PCNs to VAEs \cite{kingma_auto-encoding_2014} in this domain, but comprehensive studies are lacking, as are comparisons with other widely used models such as GANs \cite{goodfellow_generative_2014}, flow-based models \cite{rezende_variational_2015} or diffusion models \cite{sohl-dickstein_deep_2015, ho_denoising_2020}.}

Generation tasks have been considered in \cite{ororbia_neural_2022, ororbia_convolutional_2023, zahid_curvature-sensitive_2023, zahid_sample_2024}.\footnote{\cite{ororbia_neural_2022, ororbia_convolutional_2023} used a set of models dubbed {`neural generative coding'}, which is equal in spirit to PCNs as presented in this work, but with differences in notation, terminology, and minor structural details
\cite{salvatori_brain-inspired_2023}.}
Looking at \textit{sampling ability}, when measured by log-likelihood, \cite{ororbia_neural_2022, ororbia_convolutional_2023} find their models to be competitive with VAEs and GANs. This is in line with \cite{zahid_curvature-sensitive_2023, zahid_sample_2024}, which studied sampling ability using a more extensive set of metrics (Fréchet Inception Distance (FID) score, diversity, and coverage). 
Impressively, by using Langevin sampling, \cite{zahid_sample_2024} get results that match or exceed VAEs on these metrics.

In terms of computational efficiency, \cite{ororbia_neural_2022, ororbia_convolutional_2023} find that (similar to discriminative PCNs) their models also converge faster than VAEs and GANs when measured as number of epochs -- but training each epoch also takes longer due to the inference phase. Similar limitations are mentioned in \cite{zahid_predictive_2023, zahid_sample_2024}. {In sum, although there are some encouraging results, the number of studies is severely limited. Experiments on larger datasets, as well as larger models, remain to be done. For example, whether generative PCNs suffer from the same scaling issues as discriminative PCNs remains unstudied.}






\subsection{PC Graphs}\label{sec:PCgraphs}
The previous section showed how the structure of discriminative PCNs, could be extended by changing the definition of local prediction from $\bm{\mu}^\ell=f(\bm{w}^{\ell-1}\bm{a}^{\ell-1})$ to $\bm{\mu}^\ell=f(\bm{w}^{\ell+1}\bm{a}^{\ell+1})$. This can be taken one step further: PCNs can be naturally generalized to {arbitrary graphs}, called \textit{PC graphs} by \cite{salvatori_learning_2022}.\footnote{Note that PC graphs are distinct from graph neural networks (GNNs), which are used for graph-structured data \cite{ward_practical_2022}.} These are trained using IL, but dispense with the hierarchical structure of layers. {Although PC graphs have only been studied empirically by two works \cite{salvatori_learning_2022, salvatori_predictive_2024}, a recent theoretical result shows they form a superset of both discriminative and generative PCNs \cite{van_zwol_predictive_2024}. This makes them theoretically appealing, and an important avenue for further work.}
\subsubsection{Structure}
PC graphs are defined by a collection of $N$ activation {nodes} $\{a_i\}_{i=1}^N$, and error {nodes} $\{\epsilon_i\}_{i=1}^N$, with $a_i,\,\epsilon_i\in\mathbb{R}$, and $\epsilon_i=a_i-\mu_i$. The local prediction is defined as:
$$
\mu_i = \sum_{j\neq i} f(w_{ij} a_j),
$$
where $w_{ij}\in\mathbb R$, and the sum is over all the other nodes, meaning self-connections are left out ($w_{ii}=0$), as in \cite{salvatori_learning_2022}. This defines a \textit{fully-connected} PC graph. If we choose to include self-connections one simply takes the sum over all values of $i$. In vector notation, one can write $\bm{\mu} = f(\bm{wa})$. 

\begin{figure}[h!]
  \centering
  \includegraphics[width=0.55\textwidth]{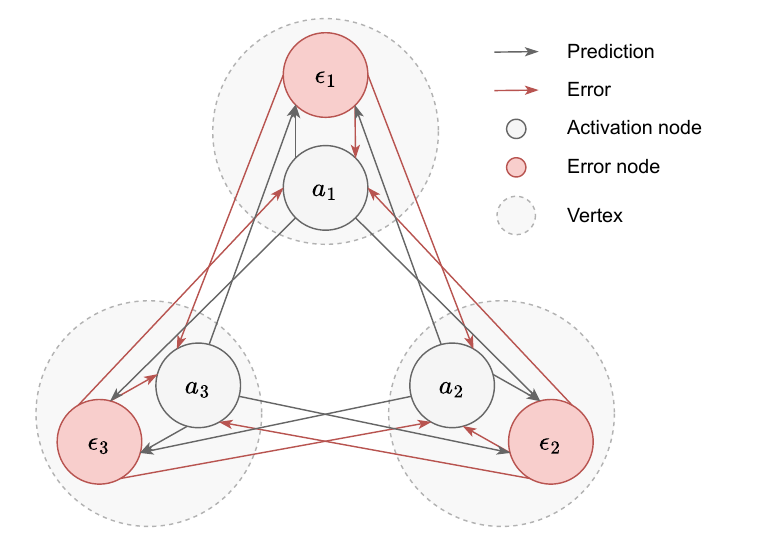}
  \caption{\textbf{Schematic architecture of a PC graph}. Three vertices are shown in a fully connected graph without self-connections. As before, predictions are shown as gray arrows, and error signals as red arrows. When training a PC graph, a subset of nodes is chosen for the data. If trained in supervised mode, an additional subset is chosen for the label.}
  \label{fig:PCG}
\end{figure}

\begin{figure}[h!]
  \centering
  \includegraphics[width=\textwidth]{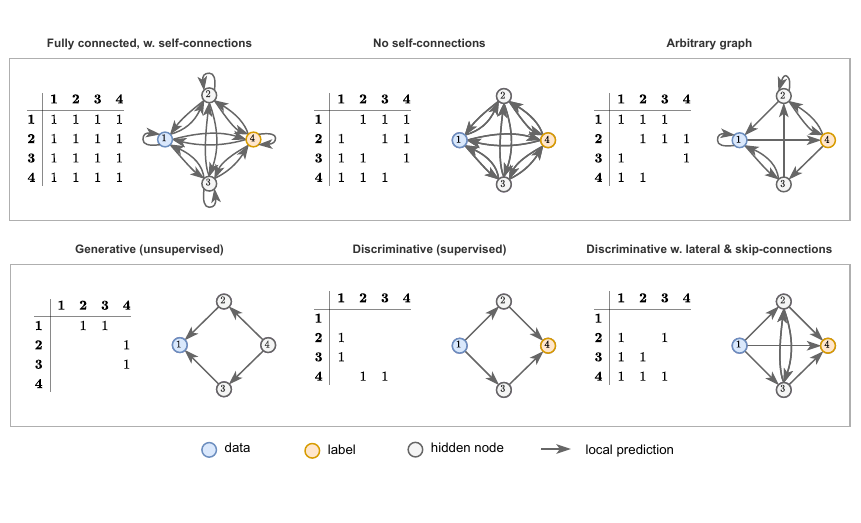}
  \caption{\textbf{Mask/adjacency matrices for a 4-node PC graph}. Different matrices lead to different architectures, like the discriminative PCN, generative PCN, or an arbitrary topology. Larger networks are obtained in the same way. See also fig. 4 in \cite{salvatori_learning_2022}.\label{fig:masking}}
\end{figure}

A small PC graph is illustrated in fig. \ref{fig:PCG}, where each activation node $a_i$ and error node $\epsilon_i$ has been grouped in a \textit{vertex} $v_i$. Importantly, graphs with different connectivity/topology can be obtained by multiplying the weight matrix with a \textit{mask} or \textit{adjacency matrix}. In this way, one obtains the architectures discussed in earlier sections: the discriminative PCN, as well as the generative PCN, depending on which weights are masked, cf. fig. \ref{fig:masking}. 

\subsubsection{Training Procedure}
Depending on the task, different nodes in the PC graph can be clamped during training. For instance, in a standard supervised mode,  the data $\{x_i\}_{i=1}^{K}$ (with dimensionality $K$ ) is clamped to a subset of activation nodes $\{a_i^x\}_{i=1}^K$, and the label $\{y_i\}_{i=1}^{M}$ (dimensionality $M$) is clamped to a second subset $\{a_i^y\}_{i=1}^M$. With simplified notation we can write this as:
\begin{equation}
	\begin{aligned}
		\{x_i\} &= \{a_i^x\} \subset \{a_i\},\quad\{y_i\} = \{a_i^y\} \subset \{a_i\}~.
	\end{aligned}
\end{equation}
For unsupervised learning, only data is clamped: $\{x_i\} = \{a_i^x\} \subset\{a_i\}$. Following this, IL can be used like earlier, where the key computations change only slightly. For example, the energy function becomes $E = \frac{1}{2}\sum_i(\epsilon_i)^2 = \frac{1}{2} \bm{\epsilon}^T\bm{\epsilon}$; other equations are shown in table \ref{tab:PCG_Vs_PCN} in appendix \ref{sec:comparisons}. Observe that the equations are obtained by simply omitting the layer $\ell$ in the equations of section \ref{sec:discriminative_PCNs}. 


\subsubsection{Testing Procedure}
Depending on how the network was trained and the desired application, different nodes can be clamped also during testing. After supervised training, classification/generation of synthetic datapoints can be done by clamping a datapoint/label respectively, running inference until convergence, and considering the produced output (a label/datapoint). 

\subsubsection{Empirical Results}
{As mentioned above, PC graphs have been studied empirically in two works \cite{salvatori_learning_2022, salvatori_predictive_2024}.} In 
\cite{salvatori_learning_2022}, models are trained for classification, generation, reconstruction, denoising, and associative memory tasks (using different adjacency matrices). An interesting result is that for classification using a \textit{fully connected} model, PC graphs perform much better than other fully connected architectures such as Boltzmann machines and Hopfield networks, up to 30\% better on MNIST. At the same time, the fully connected model does not perform comparably to hierarchical networks trained with either BP or IL. {Indeed,
\cite{salvatori_predictive_2024} further studied classification performance on PCGs improved by various tricks, finding that their best performing models use a regularizer that specifically penalizes cyclic connections.} These results might be expected, considering that depth is suggested to be critical for performance in neural network learning \cite{prince_understanding_2023}. At the same time, such statements are based mostly on empirical evidence, and lack a definitive theoretical basis. 

For the other tasks considered in \cite{salvatori_learning_2022}, only proofs of concepts are provided to demonstrate their flexibility. 
{Moreover, in \cite{salvatori_predictive_2024}, it is demonstrated in detail how PCGs may additionally be used for \textit{causal} modeling, going beyond tasks typically studied in ML.}

\subsubsection{Generalized ANNs}
At this point it is useful to compare PC graphs with PCNs discussed in section \ref{sec:ANN_to_PCN}. Fig. \ref{fig:masking} illustrated how different adjacency matrices lead to the architectures discussed in earlier sections. Given the result of \eqref{eq:discr_PCN_testing} that FNNs are equivalent to discriminative PCNs during testing, we can see them as a \textit{subset} of hierarchical PCNs, defined by the local prediction $\bm \mu^\ell = f (\bm w^{\ell\pm 1}\bm a^{\ell\pm 1})$. In turn, hierarchical PCNs can be seen as a subset of PC graphs, with a particular choice of adjacency matrix. This is illustrated in fig. \ref{fig:overview}. 

Thus, it becomes clear that formally, PCGs can be seen as types of `generalized ANNs' that go beyond hierarchical structures, by virtue of the use of IL as opposed to BP. This is very interesting, for at least two reasons. First, as observed by \cite{salvatori_learning_2022}, PC graphs allow one to train non-hierarchical structures with a brain-like topology. Speculatively, such networks could, if better understood, share some of the advantages that biological brains have over ANNs, such as vastly superior energy efficiency and parallelization. Second, from a very general perspective, topological considerations have strongly contributed to several advances in machine learning in the past. A prominent example is that of residual networks and skip connections \cite{zagoruyko_wide_2016}, which have allowed training of much deeper networks, and can improve performance on a variety of tasks \cite{prince_understanding_2023}. In a sense, the very notion of \textit{depth} in deep learning is a topological feature. As such, we consider this an important avenue for further work.

\section{PCNs as Probabilistic Latent Variable Models}\label{sec:probmodel}

Section \ref{sec:ANN_to_PCN} introduced the PCN as a type of generalized ANN. In this section we discuss a second, complementary, perspective: PCNs as \textit{probabilistic latent variable models}. Conceptualizing PCNs in this way allows a principled derivation of equations in the previous section, and brings to light connections to other methods well-known in machine learning (e.g., VAEs and linear factor models). It also reveals a number of assumptions and modeling choices, providing a basis for improvements and possible future developments.

We start our step-by-step derivation from maximum likelihood estimation in probabilistic (Bayesian) models, discussing how expectation maximization (EM) implements maximum likelihood for models with latent variables, and how different generative models lead to PC as formulated by \cite{rao_predictive_1999} or to multi-layer PC as described in the previous section. {For more mathematical background, see appendix \ref{sec:app_Bayes}.}

\subsection{Expectation Maximization}
Given data $\bm x\in \mathbb{R}^{n_x}$ (an observed variable), EM is a general method for maximizing the (log-)likelihood in models with latent variables $\bm z \in \mathbb{R}^{n_z}$, or equivalently, minimizing the negative log-likelihood, 
\begin{equation}
    \operatorname{NLL}(\theta)=-\ln p_\theta(\bm x),
\label{eq:NLL}
\end{equation}
where $\theta$ are the parameters that define the model. For some joint distribution $p_\theta(\bm x, \bm z)$, also called the generative model, it can be shown that the following two steps will minimize the NLL (see appendix \ref{sec:app_Bayes}):
\begin{align}
    \widetilde{\bm z} &= \underset{\bm z}{\operatorname{argmax}}\,p_\theta(\bm x,\bm z)&& \textbf{(E-step)},    \label{eq:E-step2}\\
\hat{\theta}&=\underset{\theta}{\operatorname{argmax}}\, p_\theta(\bm x, \widetilde{\bm z}) && \textbf{(M-step)}. \label{eq:M-step2}
\end{align}
Here, $\ln p_\theta(\bm x, \bm z)$ is also called \textit{complete data log-likelihood}, since it represents the likelihood function calculated using the full set of observed and latent variables -- to distinguish it from the likelihood calculated only with observed data \cite{bishop_pattern_2006}. We will label the \textit{negative complete data log-likelihood} as $E$: 
\begin{equation}
	E(\bm x, \bm z)=-\ln p_\theta(\bm x, \bm z)
	\label{eq:PCE}
\end{equation}
which refers to \textit{energy}, as in section \ref{sec:ANN_to_PCN} (cf. section \ref{sec:ebms} for discussion on energy-based models). We will choose to minimize $E$ as opposed to maximizing $\ln p_\theta(\bm x, \bm z)$.  It has furthermore been shown (cf. appendix \ref{sec:app_Bayes}, ref. \cite{neal_view_1998}) that \textit{partial} E-steps and M-steps will also minimize the NLL: 
\begin{align}
\Delta \bm{z} &\propto  -\frac{\partial E}{\partial\bm{z}}&& \textbf{(partial E-step)},  \label{eq:E-step_EBM}  \\
\Delta {\theta} &\propto -\frac{\partial E}{\partial {\theta}}&& \label{eq:M-step_EBM}\textbf{(partial M-step)}.
\end{align}
This, with a change of notation, is \textit{incremental IL} as presented in section \ref{sec:ANN_to_PCN}. To obtain \textit{standard IL}, one employs a full E-step, followed by a partial M-step. 

{Understanding why EM minimizes the NLL can be done by considering a specific KL-divergence:
\begin{equation}
    \begin{aligned}
\mathcal{F}&\equiv D_{\text{KL}}[q_\phi(\bm z)||p_\theta(\bm x,\bm z)]\\
&=D_{\text{KL}}[q_\phi(\bm z)||p_\theta(\bm z|\bm x)] -\ln p_\theta(\bm x)~,
\end{aligned}
\end{equation}
where $\mathcal{F}$ is called the variational free energy, and $q_\phi(\bm z)$ is a function that depends on parameters $\phi$. Crucially, since the KL divergence is non-negative, one has
\begin{equation}
    \mathcal{F} \geq - \ln p_\theta(\bm x)~,
\end{equation}
i.e., $\mathcal{F}$ is an {upper bound} of the NLL.\footnote{One can equivalently write $-\mathcal{F} \leq \ln p_\theta(\bm x)$, i.e. $-\mathcal{F}$ as a lower bound to $\ln p_\theta(\bm x)$. The marginal log-likelihood is also called the \textit{model evidence}, hence $-\mathcal{F}$ is also called the evidence lower bound, or \textit{ELBO} \cite{prince_understanding_2023}.} Specifically, under certain assumptions (cf. appendix \ref{sec:app_Bayes}), one finds
\begin{equation}
    \mathcal{F} \approx E~.
    \label{eq:FapproxE}
\end{equation}
For a given generative model, this can be tractably minimized, with the result of minimizing the NLL. Minimization with respect to $\phi$ and $\theta$ can be shown to equate to the E-step and M-step as defined above.} {This discussion relates closely to the topic of \textit{variational inference}. Compared to most works in the PC literature, our works takes a non-standard approach to treating this topic, which we justify below.} 

\subsection{Variational Inference}
{
In variational inference the goal for $q_\phi(\bm z)$ is to approximate the posterior $p_\theta(\bm z| \bm x)$, which is intractable ($q$ is then appropriately called the variational posterior). Hence, the functional 
$$D_{\text{KL}}[q_\phi(\bm z)||p_\theta(\bm z|\bm x)]$$
is also intractable, but since $-\ln p_\theta(\bm x)\geq 0$, one has 
$$
\mathcal{F} \geq D_{\text{KL}}[q_\phi(\bm z)||p_\theta(\bm z|\bm x)]~.
$$ 
Therefore, $\mathcal{F}$ is an upper bound not only of the NLL but also of this functional. Hence, the aforementioned minimization also has the effect of optimizing $\phi$ to make $q$ better match the true posterior: variational inference.}

{Thus it is clear that PC is closely {related} to VI. However, the two are often \textit{equated} in the literature \cite{millidge_predictive_2022, pinchetti_predictive_2022}, which we argue below is somewhat misleading given how the term VI is typically used within ML \cite{mackay_information_2003, bishop_pattern_2006, murphy_probabilistic_2023}. Moreover, although EM is mentioned in tandem with PC much less often than VI, we argue it is the essence of the algorithm---its two steps are equivalent to the inference and learning steps in PC. For these two reasons, unlike most works in the PC literature, we defer to appendix \ref{sec:app_Bayes} the derivations involving the variational free energy, emphasizing the role of EM here.}

{Our argument is that in the derivation leading up to \eqref{eq:FapproxE} (cf. appendix \ref{sec:app_Bayes}), one assumes for $q_\phi(\bm z)$ either a delta function or a Gaussian (a so-called Laplace approximation). In the former case, $q$ reduces directly to a point estimate, i.e. MAP estimation. In the latter case one initially takes into account the second moment of $q$, but this is disregarded in practice \cite{whittington_approximation_2017, millidge_predictive_2022, innocenti_only_2024} (with the exception of \cite{zahid_curvature-sensitive_2023}). This is categorically unlike, e.g., variational autoencoders, where the uncertainty (i.e., second moment) in $q$ is explicitly accounted for \cite{prince_understanding_2023}. Indeed, seminal ML textbooks \cite{mackay_information_2003, bishop_pattern_2006, murphy_probabilistic_2023} introduce VI as explicitly going \textit{beyond} MAP and the Laplace approximation. Thus, calling PC a form of VI is somewhat confusing in that it ignores higher moments in practice; it equates to VI only in the most limited sense. }

\subsection{Generative Models}
Having described the general process for minimizing the NLL, we can now proceed with modeling $p_\theta(\bm x, \bm z)$.
\subsubsection{PCNs of Rao \& Ballard}
\begin{figure}[ ]
  \centering
  \includegraphics[width=0.3\textwidth]{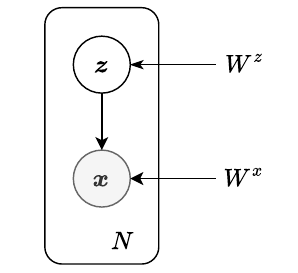}
  \caption{\textbf{Directed graphical model for PCNs of Rao \& Ballard} \cite{rao_predictive_1999}, with variances set to unity. Probabilistic graphical models represent how a joint probability distribution of random variables (nodes) can be factorized, here showing \eqref{eq:decomPC} with $\theta=\{W_z, W_z\}$ representing deterministic parameters. The graphical model is surrounded by a \textit{plate} labelled with $N$ indicating that there are $N$ nodes of this kind, one for each datapoint (cf. e.g.\cite{bishop_pattern_2006} ).}
  \label{fig:gm_rao}
\end{figure}

Here we derive the model of Rao \& Ballard, i.e., PC with two layers. First factorize the generative model as:
\begin{equation}
	p_\theta(\bm x, \bm z) = p(\bm x|\bm z)p(\bm z)~.
	\label{eq:decomPC}
\end{equation}
To obtain the original results of Rao \& Ballard, one assumes that both distributions on the right-hand side may be well-approximated by Gaussians. Hence for $p(\bm x|\bm z)$ we take
\begin{equation}
	p(\bm x|\bm z)=\mathcal{N}(\bm x;\bm \mu_x,\Sigma_x)=\frac{1}{\sqrt{(2 \pi)^{n_x}\operatorname{det}\Sigma_x }}
 \exp \left( -\frac{1}{2}(\bm{x}-\bm{\mu}_x )^{T} (\Sigma_x)^{-1}(\bm{x}-\bm{\mu}_x ) \right)~,
 \label{eq:gauss1}
\end{equation}
and likewise for the prior, $ p(\bm z)=\mathcal{N}(\bm z;\bm \mu_z,\Sigma_z)~$. Furthermore, take $\bm \mu_x = f(W_x \bm z)$ and $\bm \mu_z =f(W_z \bm z_p)$ with $f, W_x, W_z$ defined as in section \ref{sec:ANN_to_PCN}, and $\bm z_p$ a parameter for the prior (either fixed or learnable). $\Sigma_x\in \mathbb{R}^{n_x^2}, \Sigma_z \in \mathbb{R}^{n_z^2}$ are covariance matrices. Thus, $\theta = \{W_x\,, W_z,\,\Sigma_x, \Sigma_z\}$ are the parameters that define the generative model. For convenience, we also define the errors $\bm \epsilon_x = \bm x - \bm \mu_x$ and $\bm \epsilon_z = \bm z - \bm \mu_z$ as in section \ref{sec:ANN_to_PCN}. The model is illustrated schematically in fig. \ref{fig:gm_rao}.

With the decomposition \eqref{eq:decomPC}, the energy \eqref{eq:PCE} is
\begin{equation}
	\begin{aligned}
		E(\bm x, \bm z)&=-\ln p(\bm x|\bm z)-\ln p(\bm z)\\
			       &= \frac{1}{2}\bigg( (\bm \epsilon_x)^{T} (\Sigma_{x})^{-1}\bm \epsilon_x + (\bm \epsilon_z)^{T} (\Sigma_{z})^{-1}\bm \epsilon_z + \ln \left[\det \Sigma_x \det \Sigma_z\right] \bigg)+\text{const.}~,
			       \label{eq:E1}
	\end{aligned}
\end{equation}
where on the second line we have inserted the Gaussian ans\"atze \eqref{eq:gauss1}, the prior, and the constant -- which is irrelevant to the optimization problem -- is $\frac{1}{2}(n_x+n_z)\ln 2\pi$. Choosing diagonal bases so that $\Sigma = \sigma^2 I$, and disregarding the constant and the factor of 1/2, we have
\begin{equation}
	E(\bm x, \bm z)=  \frac{1}{\sigma_x^2} (\bm \epsilon_x)^2 +  \frac{1}{\sigma_z^2}(\bm \epsilon_z)^2~,
	\label{eq:RBenergy}
\end{equation}
which is the energy function used by Rao \& Ballard \cite{rao_predictive_1999} (with variances taken as constants). With a model for $E$ in hand, we can now apply EM, cf. \eqref{eq:E-step2} and \eqref{eq:M-step2}.

\subsubsection{Multi-layer PC}

\begin{figure}[ ]
  \centering
  \includegraphics[width=0.33\textwidth]{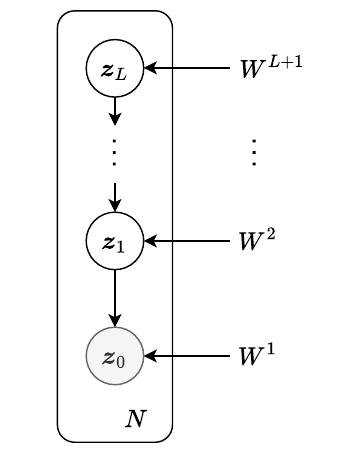}
  \caption{\textbf{Graphical model for multi-layer PCNs}, with $\Sigma^\ell=I$. This defines the factorization of the observed variable (here $\bm z^0$) and latent variables ($\bm z^\ell$), cf. \eqref{eq:markov}. $\theta = \{W^\ell\}_{\ell=0}^{L+1}$ is shown as the set of deterministic parameters, and a plate labeled with $N$ representing $N$ nodes for the datapoints.}
  \label{fig:gm_ML}
\end{figure}

We now extend the previous section to a model with $L$ layers (vectors) of latent variables $\{\bm{z}^\ell\}_{\ell=1}^L$ in a hierarchical structure.
It is convenient to label the observed variables $\bm x$ as $\bm x = \bm z^0\in\mathbb{R}^{n_\ell}$. Thus our generative model is $p_\theta(\bm{x}, \{\bm z^\ell \})=p_\theta(\bm{z}^0,...,\bm{z}^L)= p_\theta(\{\bm z^\ell \})$.
The hierarchical structure then translates to the statement that each layer is conditionally independent given the layer above, i.e., 
\begin{equation}
p_\theta(\{\bm z^\ell \})=p_\theta(\bm{z}^L)\prod_{\ell=0}^{L-1} p_\theta(\bm{z}^\ell|\bm{z}^{\ell+1})~,
\label{eq:markov}
\end{equation}
so that the energy becomes
\begin{equation}
	\begin{aligned}
 		E(\{ \bm z^\ell \})&=-\ln p_\theta(\{\bm z^\ell \})\\
 		&= -\ln p_\theta(\bm z^L)-\sum_{\ell=0}^{L-1}\ln p_\theta(\bm z^\ell|\bm z^{\ell+1}) ~.
\label{eq:E2} 
	\end{aligned}
\end{equation}
As in the simple model above, we assume a multidimensional Gaussian for the conditional distribution of each layer, as well as for the prior:\footnote{This is a reasonable assumption for traditional (fully-connected, feedforward) ANNs with large layer widths, $n_\ell\sim n\gg1$, which becomes exact in the so-called large-width limit $n\mapsto\infty$, due to the central limit theorem. However, non-Gaussianities appear at finite width (see, e.g., \cite{roberts_principles_2022, grosvenor_edge_2022} for theoretical treatments), and it is unclear how realistic this assumption is in more general networks involving intralayer or feedback connections. }
\begin{equation}
	\begin{aligned}
		p_\theta(\bm{z}^\ell|\bm{z}^{\ell+1})&=\mathcal{N}(\bm{z}^\ell;\bm \mu^\ell,\Sigma^\ell)\\
 		p(\bm z^L) &= \mathcal{N}(\bm{z}^L;\bm \mu^L,\Sigma^L)
	\end{aligned}
\end{equation}
with mean $\bm \mu^\ell = f(W^{\ell+1}\bm{z}^{\ell+1})$ and covariance matrix $\Sigma^\ell \in \mathbb{R}^{ n_\ell \times n_\ell }$, where ${W^\ell \in \mathbb{R}^{ n_\ell \times n_{\ell+1}}}$ and $\bm \mu^L=\bm f(W^{L+1}\bm z_p)$. These parameters, $\theta = \{W^\ell,\,\Sigma^\ell\}_{\ell=0}^{L+1}$,  define the generative model, which is illustrated in fig. \ref{fig:gm_ML}.
As before, we define the errors $\bm \epsilon^\ell = \bm z^\ell - \bm\mu^\ell$. We then have
\begin{equation}
-\ln p(\bm z^\ell|\bm z^{\ell+1})=\frac{1}{2}\left( (\bm \epsilon^\ell)^{T} (\Sigma^{\ell})^{-1}\bm \epsilon^\ell + \ln \det \Sigma^\ell + n_\ell\ln 2\pi \right)
\end{equation}
such that $E$ becomes
\begin{equation}
	E(\{ \bm z^\ell \})= \frac{1}{2}\sum_{\ell=0}^L\left( (\bm \epsilon^\ell)^{T} (\Sigma^{\ell})^{-1}\bm \epsilon^\ell + \ln \det \Sigma^\ell \right) + \text{const}~,
\end{equation}
where the constant is $\sum_{\ell=0}^L n_\ell\ln 2\pi$. Dropping this, and choosing a diagonal basis so that $\Sigma^{\ell}= I$, we obtain
\begin{equation}
	E(\{ \bm z^\ell \})= \frac{1}{2}\sum_{\ell=0}^L (\bm \epsilon^\ell)^2 
\label{eq:E_with_C}
\end{equation}
which is the energy of Section \ref{sec:ANN_to_PCN}; cf. also \eqref{eq:RBenergy}. Applying EM to this then yields IL as presented in section \ref{sec:ANN_to_PCN} (replacing $\bm z^\ell$ by $\bm a^\ell$\footnote{Following typical ML notation, we use $\bm z$ for latent variables in the context of graphical models, and $\bm a$ for activation neurons in the neural network context.}).

\subsection{Learning in PCNs Revisited}

\subsubsection{Discriminative and Generative PCNs}

At this point we can provide a number of interesting interpretations of earlier sections. Notice that in the derivation above, the model had observed variables at layer $\ell=0$. This corresponds to \textit{generative PCN,} as discussed in section \ref{sec:generative_PCN}. If trained in supervised mode, we can understand the clamping of the label $\bm y^{(n)}$ to the final layer $\bm z^p$ as follows: $\bm y^{(n)}$ becomes a parameter of the prior (e.g. its mean). If trained in unsupervised mode, it is treated just like the other hidden variables. Interestingly then, for a discriminative PCN, we see that the clamped data $\bm x^{(n)}$ is a parameter for the prior, and the label is now the observed variable. This is shown in table \ref{tab:PCN_direction}.

\begin{table}[h!]
\centering
\caption{\textbf{Overview of learning modes in the probabilistic perspective of PCNs.}}
\setlength{\tabcolsep}{1pt} 
\renewcommand{\arraystretch}{1.2} 
\resizebox{\columnwidth}{!}{%
\begin{tabular}{>{\bfseries}l p{3.1cm} p{3.3cm} p{3.1cm} p{3.1cm}}
\toprule
 & & \multicolumn{2}{c}{\textbf{Generative PCN}} & \textbf{Discriminative PCN} \\
\cmidrule(lr){3-4} \cmidrule(lr){5-5}
 & & Unsupervised learning & Supervised learning & Supervised learning \\
\midrule
$\bm z^L$ & Prior parameter & - & Label $\bm y^{(n)}$ & Data $\bm x^{(n)}$ \\
$\bm z^0$ & Observed variable & Data $\bm x^{(n)}$ & Data $\bm x^{(n)}$ & Label $\bm y^{(n)}$ \\
\bottomrule
\end{tabular}
}
\label{tab:PCN_direction}
\end{table}
\subsubsection{Precision Matrices}\label{sec:precision}
In the context of predictive coding, the covariance matrices $\Sigma^\ell$ are often interpreted as the uncertainty at each layer around the mean $\bm \mu^\ell$, with their inverse $(\Sigma^\ell)^{-1}\eqqcolon\Pi^\ell$ sometimes called \emph{precision matrices} \cite{friston_learning_2003, millidge_theoretical_2023, marino_predictive_2022}. Within neuroscience and psychology, they have been used in a wide range of models \cite{millidge_predictive_2022}, where it has been suggested that they implement a type of attention \cite{feldman_attention_2010}. By up-weighting the error, increasing the precision of a variable would serve as a form of gain modulation. From the ML perspective, instead of setting them to identity matrices, one can add minimization of $E$ w.r.t. $\Pi^\ell$ as an additional part of the M-step, seeing them simply as additional parameters of the generative model. However, their practical utility remains unclear, with varying results. In particular, \cite{ofner_generalized_2022} do not find advantages of additionally learning for generative modeling tasks, while \cite{ororbia_neural_2022} do appear to find some benefit, suggesting they act as a type of lateral modulation that induces a useful form of sparsity in the model. Comparisons with VAEs are made in \cite{pinchetti_predictive_2022}, with a different theoretical interpretation by \cite{millidge_theoretical_2023} (cf. section \ref{sec:learning} for further discussion).

\subsection{Connections to Other Latent Variable Models}
PCNs share various properties with other probabilistic models in machine learning. We briefly discuss some of these connections here, leaving more extensive comparisons (e.g., on training methods, sampling) for future work.
\subsubsection{Linear Factor Models.}
\begin{figure}[h!]
  \centering
  \includegraphics[width=0.85\textwidth]{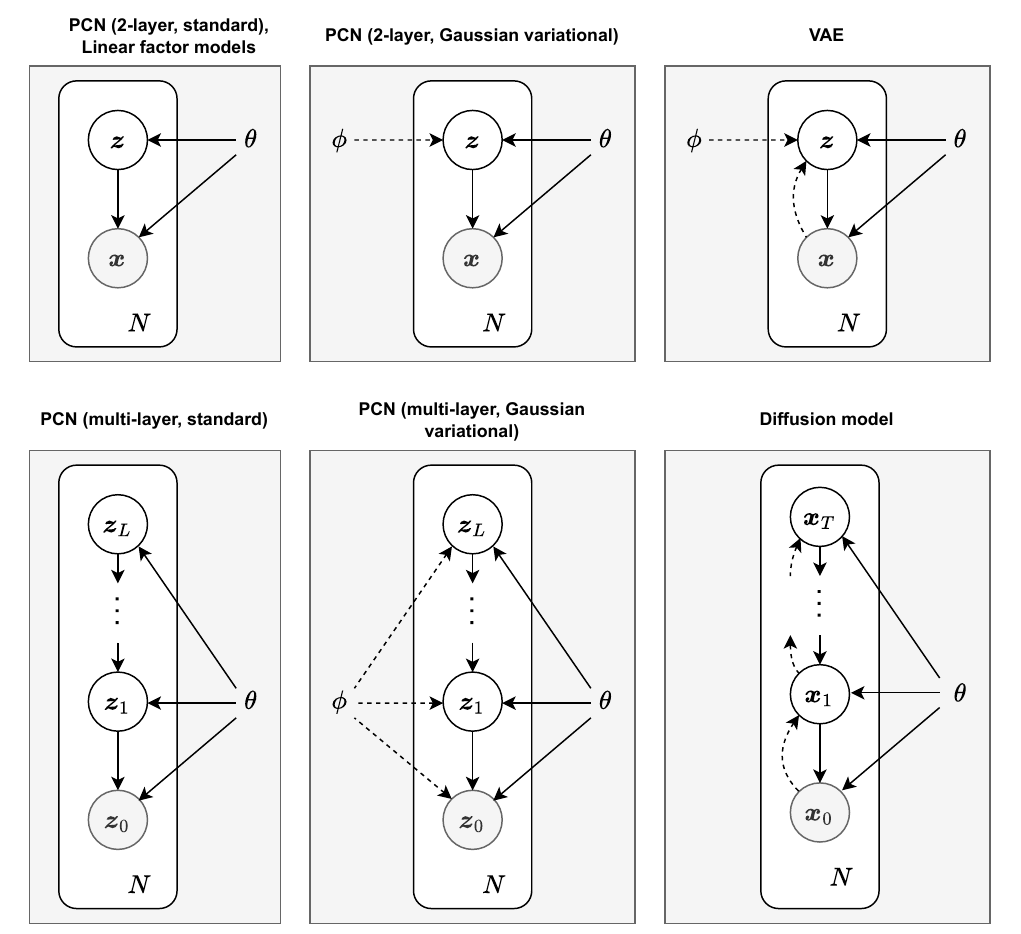}
  \caption{\textbf{Graphical model comparison of PCNs and other probabilistic models}, with $\theta$ model parameters and $\phi$ variational parameters (cf. appendix \ref{sec:app_Bayes}). Standard PCNs (left column) use MAP inference of the latent variables during the E-step, without variational parameters. Using a Gaussian variational (middle column) adds the mean and variance of the variational as parameters $\phi$ which may additionally be varied, shown by dotted lines. This is similar to $\phi$ in a VAE (top right). VAEs differ from PCNs in that their variational posterior is conditioned explicitly on $\bm x$ (i.e. $q(\bm z|\bm x)$ in VAEs vs. $q(\bm z)$ in PCNs). Diffusion models similarly use a conditioned variational, and are similar to hierarchical VAEs \cite{prince_understanding_2023}. However, diffusion models do not use variational parameters $\phi$, and their latent variables have the same dimensionality as the observed variable, hence their use of $\bm x$ instead of $\bm z$.\label{fig:gm_comp}}
\end{figure}

With $f$ as a linear function, the two-layer PCN model discussed above is equivalent to a \textit{linear factor model}. This is illustrated in fig. \ref{fig:gm_comp}. Specifically, if we take $\bm \mu_z = \bm 0$, $\Sigma_z = I$, and $\Sigma_x=\operatorname{diag}\bm{\sigma}^2$ with $\bm \sigma^2=[\sigma_1^2, \sigma_2^2,...,\sigma_n^2]^T$, then we obtain exactly \textit{factor analysis}. If additionally take $\Sigma_x=\sigma^2 I$, i.e. the variances $\sigma_i^2$ are all equal to each other, we obtain \textit{probabilistic PCA} \cite{goodfellow_deep_2016}. Like in PCNs, parameters in this model can be learned using the EM algorithm \cite{bishop_pattern_2006}. 

\subsubsection{Variational Autoencoders and Diffusion Models}
VAEs share many properties with PCNs, as discussed in \cite{marino_predictive_2019, marino_predictive_2022, ofner_generalized_2022, pinchetti_predictive_2022}. They make use of a similar graphical model as two-layer PCNs (and linear factor models). VAEs are trained using variational free energy as an upper bound for the NLL (cf. appendix \ref{sec:app_Bayes}). While they use different optimization techniques, these can be seen as design choices for solving the same inference problem \cite{marino_predictive_2022}. 
A number of key differences can be identified. First, the conditional $p(\bm x|\bm z)$ is parametrized by a linear function, followed by a nonlinearity (i.e.$f(W^z\bm z)$). In contrast, VAEs use a deep neural network (trained with BP). Moreover, in the variational inference perspective, whereas PCNs use a variational posterior of the form $q(\bm z)$, VAEs use $q_\phi(\bm z|\bm x)$, i.e. with an explicit conditioning on data (visualized by dotted lines in fig. \ref{fig:gm_comp}). This is parameterized by a separate neural network through parameters $\phi$, which are learned jointly with $\theta$ using BP. For further comparisons, see \cite{marino_predictive_2019, marino_predictive_2022}.

Adding multiple layers to VAEs results in \textit{hierarchical VAEs}, which can similarly be compared to multi-layer PCNs. \textit{Diffusion models} \cite{sohl-dickstein_deep_2015, ho_denoising_2020} are also shown in fig. \ref{fig:gm_comp}, which share (most of) their graphical structure with hierarchical VAEs. Like VAEs, diffusion models factorize the variational posterior using conditional distributions, i.e. $q(\bm x^\ell|\bm x^{\ell+1})$, but this is not parametrized by separate parameters $\phi$. I.e. the variational model is not learned, but fixed.

In terms of performance, comparisons between multi-layer PCNs and (non-hierarchical) VAEs have been studied by \cite{ororbia_neural_2022, ofner_generalized_2022, pinchetti_predictive_2022, zahid_sample_2024}. PCNs seem to compare favorably, with similar or improved results on several metrics, especially in the most recent work of \cite{zahid_sample_2024}. At the same time (similar to the picture that emerged for PCNs used for supervised learning), PCNs seem to be more computationally costly to train, though again relatively little work has been done on training optimization in comparison to that for VAEs. Comparisons with hierarchical VAEs and diffusion models have not yet been done to the authors' knowledge.

\subsubsection{Energy-Based Models}\label{sec:ebms}
PCNs are often cited to belong to the class of Energy-Based Models (EBMs) \cite{millidge_theoretical_2023, millidge_backpropagation_2023, song_inferring_2024}. However, in these works the term `EBM' is used in a different way than is typical in ML. The typical use of EBM  \cite{lecun_tutorial_2006, tomczak_deep_2022} is based on the observation that any probability distribution function can be parametrized by:
\begin{equation}
	p_\theta(\bm x)=\frac{\exp \left(-E_\theta(\bm x)\right)}{Z(\theta)}~,
 \label{eq:EBM}
\end{equation}
where $E_\theta(\bm x)$ is the \textit{energy} function, and $Z(\theta)$ is the partition function, i.e., the sum of all possible states. This use comes from physics where \eqref{eq:EBM} is called a Boltzmann distribution, corresponding to a system at thermal equilibrium described by a canonical ensemble. The prototypical example of EBMs are Boltzmann machines, which have $E_\theta(\bm x) = -\frac{1}{2}\bm x^T W \bm x$ \cite{ackley_learning_1985, mackay_information_2003}. 

In contrast, in \cite{millidge_theoretical_2023, millidge_backpropagation_2023, song_inferring_2024} an EBM is instead defined through \eqref{eq:E-step_EBM} and \eqref{eq:M-step_EBM} \cite{millidge_theoretical_2023, millidge_backpropagation_2023, song_inferring_2024}. This definition has also been used by \cite{scellier_equilibrium_2017}, and encompasses continuous Hopfield networks and networks trainable by Equilibrium Propagation. Indeed, PCNs have much in common with these techniques, cf. \cite{millidge_backpropagation_2023} and section \ref{sec:learning}. 

Regarding EBMs as defined by \eqref{eq:EBM}, comparisons of PCNs with this class of models has not yet been done to the authors' knowledge. We briefly mention some differences here, comparing to Boltzmann machines.\footnote{Boltzmann machines are characterized by all-to-all connectivity between nodes. Another important class of models is obtained by restricting the weight matrix to only include connections between $\bm x$ and $\bm z$, whereupon one obtains restricted Boltzmann machines (RBMs), which can be stacked to obtain a hierarchical structure with interesting Bayesian interpretations \cite{goodfellow_deep_2016}.}  If one includes latent variables $\bm z$, one may write $p_\theta(\bm x)=\sum_{\bm z} p_\theta(\bm x, \bm z)$ (with a sum instead of an integral because nodes in Boltzmann machines are typically discrete), with \cite{mackay_information_2003}:
\begin{equation}
	p(\bm x, \bm z) = \frac{\exp \left(-E_\theta(\bm x, \bm z)\right)}{Z(\theta)}~.
 \label{eq:EBM2}
\end{equation}
Now, observe that the generative model of PCNs, by definition of $E$ (cf. \eqref{eq:PCE}), can be written as $p_\theta(\bm x, \bm z) = \exp(-E_\theta(\bm x, \bm z))$, which is similar to \eqref{eq:EBM2}. However, note that $E$ in this equation is not a proper `energy' as in \eqref{eq:EBM} due to the lack of normalization in $Z$. 
Comparing EBMs to PCNs, one may first observe that $E$ is of course different. Another difference is that PCNs minimize $-\ln p_\theta (\bm x, \bm z)$, whereas EBMs typically minimize the NLL itself -- not by using EM, but taking derivatives of the NLL w.r.t. model parameters, calculated using MCMC techniques, e.g. Gibbs sampling \cite{goodfellow_deep_2016}. Clarifying these differences further could be an interesting direction for future work, further bringing PC and ML closer together.

\section{PCNs and Inference Learning}\label{sec:learning}
This section discusses IL in detail, which has been the focus of much of the recent literature -- particularly on the connection between IL and BP. Various ways of modifying IL have been introduced to approximate or replicate BP's weight updates. More recent works however, instead consider IL in its `natural regime' -- meaning without modifications as presented in section \ref{sec:ANN_to_PCN}. It has also become understood that PCNs/IL share properties with other algorithms, like target propagation \cite{millidge_theoretical_2023} and trust region methods \cite{innocenti_understanding_2023}. We review these below. In general, much is still unclear about the properties of PCNs and IL. In particular, the properties of IL in generative PCNs and PC graphs remain underexplored.

\subsection{Inference Learning and Backpropagation}\label{sec:BP-relation}
As far as we are aware, the first use of PC applied to an ML task was \cite{whittington_approximation_2017}. They showed, under certain strict conditions, that the weight updates of IL on discriminative PCNs approximates the parameter updates of BP in  FNNs. Subsequently, this result was shown to generalize -- again with strict conditions -- to any computational graph (a decomposition of complex functions such as DL architectures, e.g. FNNs, CNNs and RNNs, into elementary functions ) \cite{millidge_predictive_2022, rosenbaum_relationship_2022}. Using a different method, it was shown that a variation of PC, called Z-IL, gives the \textit{exact} weight updates of BP on any computational graph \cite{song_can_2020, salvatori_predictive_2021, salvatori_reverse_2022}. The reason this works is that Z-IL uses a highly specific number of inference steps that matches the number of layers, and updates weights of different layers at very specific steps during inference. Another requirement is that the `feedforward pass' initialization of hidden nodes is done (cf. section \ref{sec:initialization}). (Alg. \ref{alg:ZIL} in appendix \ref{sec:comparisons} shows pseudocode for Z-IL.)

\subsection{Inference Learning and Other Algorithms}
Since it has been known that other algorithms (Equilibrium Propagation \cite{scellier_equilibrium_2017}, Contrastive Hebbian Learning \cite{xie_equivalence_2003}) also approximate BP, work by \cite{millidge_backpropagation_2023} showed that these various approximations can be understood as a consequence of certain properties of energy-based models (EBMs), as defined by \eqref{eq:E-step_EBM} and \eqref{eq:M-step_EBM} (cf. section \ref{sec:ebms}).

A practical limitation of these works is that the various approximations and variants of PC always remain slower than the (highly optimized) BP \cite{zahid_predictive_2023}. As such, they do not provide a clear argument for using e.g. Z-IL in practical applications.

Another interesting theoretical connection found by \cite{millidge_theoretical_2023} is that PC can be understood to \textit{interpolate} between BP and a third algorithm for training neural networks, called \textit{target propagation} (TP) \cite{lee_difference_2015}. This algorithm propagates \textit{target activations} (not errors) to the hidden layers of the network, followed by updating weights of each layer to move closer to the target activation. The authors show that{, for linear networks,} if a PCN is trained \textit{with only labels clamped}, the other activities converge to the targets computed by TP. At the same time (as was shown in section \ref{sec:ANN_to_PCN}),\textit{with only data clamped}, the activations computed with IL are equal to feedforward ANN activations. As such, the equilibrium of IL's inference phase {with \textit{both} data and labels clamped} can be interpreted as an `average' {, an interpolation} of the feedforward pass values of the network, and the local targets computed by TP. {Although this result is shown rigorously only for linear networks, this intuition is shown to hold for non-linear networks using experiments.} Furthermore, the authors observe that \textit{precision weights} $\Pi^\ell=(\Sigma^\ell)^{-1}$ provide a means to change the relative weighting of the feedforward and feedback influences on the network dynamics (cf. section \ref{sec:precision}). {Thus, IL is in some sense a more general learning algorithm than BP and TP, by unifying some of their properties.}

\subsection{Inference Learning in its Natural Regime}

Following insights from the aforementioned references {\cite{song_can_2020, salvatori_predictive_2021, millidge_predictive_2022, rosenbaum_relationship_2022, salvatori_reverse_2022, zahid_predictive_2023, millidge_backpropagation_2023, millidge_theoretical_2023}}, recent works have instead started to consider the `natural' regime of IL as presented in section \ref{sec:ANN_to_PCN}, without modifications. So far, no comprehensive account of the properties of IL exists, so we review the somewhat disparate theoretical results in the literature.

Although most works study properties of the learning algorithm as a whole, the work of  \cite{frieder_non-convergence_2022} showed a useful result concerning the inference phase. Using a dynamical systems perspective, they formally prove that the {inference} phase (during training and testing) converges only when $\gamma$ and $\alpha$ (the inference and learning rate, respectively) are smaller than 1. This is always done in practice, but \cite{frieder_non-convergence_2022} gave theoretical assurance that this is indeed the case.
Then, the first contribution to a theory for why IL sometimes performs better than BP was provided by \cite{song_inferring_2024}, which according to the authors lies in \textit{reduced weight interference}. `Catastrophic interference' is a well-known pathology of BP to abruptly and drastically forget previously learned information upon learning new information \cite{kirkpatrick_overcoming_2017}. According to \cite{song_inferring_2024}, this is reduced in IL by using a mechanism called \textit{prospective configuration} (cf. box below). According to \cite{song_inferring_2024} this helps explain improvements observed in continual learning tasks (which are a challenge precisely due to this interference), online learning, and convergence.

\begin{mdframed}[backgroundcolor=lightergray]
\subsubsection*{Prospective Configuration}\label{sec:prosp_config}

{Prospective configuration provides a complementary perspective on the difference between BP and IL, relating to the importance of activation changes in IL. Note that since the output is clamped in IL, activation changes are {`guided' by the target}. This is not the case in BP (which only uses information from the target during the weight update). As a consequence, at IL's equilibrium, activations take values they ideally {should have}, in order to correctly classify the input according to the provided target. }

{Also consider the activations after a single datapoint is processed in an FNN vs. a PCN. In the former, activations contain nothing but noise by virtue of random weight initializations. Only when processing the \textit{second} datapoint will information from the target have transferred to the activations, as a consequence of the first weight update. Hence in BP, \textit{activation changes follow weight changes}. By contrast, in IL the weight updates consolidate the pattern of activations at inference equilibrium: \textit{weight changes follow activation changes}. In this way, activation changes `foresee' the weight changes, making them \textit{prospective}. }
\end{mdframed}

In a parallel line of work, \cite{alonso_theoretical_2022} showed both theoretically and empirically that IL approximates \textit{implicit SGD},  

which provides increased stability across learning rates. The authors also develop a variant of PC in which updates become equal to implicit SGD, further improving the stability. This is extended in \cite{alonso_understanding_2024}, where the authors observe that implicit SGD is sensitive to {second-order information}, i.e. the second derivative of the loss landscape (its curvature). This can speed up convergence. As such, when IL approximates implicit SGD, IL uses this information, which explains the faster convergence also observed by \cite{song_inferring_2024}. {A complementary perspective on this comes from \cite{mali_tight_2024, ishikawa_local_2024}, showing that IL approximates \textit{quasi-Newton} methods, which use second-order information explicitly.}

This connects to \cite{innocenti_understanding_2023}, which related IL to \textit{trust region (TR) methods}. These are a well-known method in optimization, which can be contrasted with line-search methods such as gradient descent \cite{yuan_recent_2015}. Such methods define a neighborhood around the current best solution as a \textit{trust region} in each step, which is changed adaptively in addition to the step direction \cite{yuan_recent_2015}. The authors show \cite{innocenti_understanding_2023} that IL's inference phase of PC can be understood as solving a TR problem on the BP loss,  using a trust region defined by a second derivative of the energy. By using this second-order information, trust region methods are well-known to be better at escaping saddle points (non-extremal points with zero gradient). Indeed, \cite{innocenti_understanding_2023} show that these properties transfer to IL, thus drawing a similar conclusion as \cite{alonso_understanding_2024}. They prove on toy problems that IL escapes saddle points faster, and provide evidence that this happens in larger networks by training deep chains, where IL converges significantly faster than BP. {This analysis is extended in \cite{innocenti_only_2024}, where the authors prove that many saddle points in the loss landscape become \textit{strict} at the inference equilibrium. This follows from the observation that, for linear networks, the PC energy is a non-trivial rescaling of the MSE loss. This helps them formalize the observations of \cite{innocenti_understanding_2023}, since non-strict saddles can slow down learning exponentially for (stochastic) gradient descent. Their results lead to the conjecture that inference makes \textit{all} saddles strict, thus making the loss landscape more benign and robust to vanishing gradients. These results help to explain the empirical results showing faster convergence of PC compared to BP (cf. section \ref{sec:empirical1}). }

Theoretical results of \cite{millidge_theoretical_2023} are concerned with the convergence in PCN's energy landscape. The authors showed that when activity nodes are initialized with the so-called {`energy gradient bound'}, IL converges to \textit{the same extrema} in the loss as BP. This can be understood by writing PCN's energy as a sum of two terms $E = \mathcal{L}+\widetilde{E}$, where $\mathcal{L}= (\bm \epsilon^L)^2$ is the energy of the last layer, equal to the BP MSE loss, and $\widetilde{E}=\sum_{\ell=1}^{L-1}(\bm\epsilon^\ell)^2$ is the residual energy. Now, during inference, minimizing $E$ is guaranteed to minimize $\mathcal{L}$ if the following holds that (cf. \cite{millidge_theoretical_2023} for a derivation) $-\left(\pdv{\mathcal{L}}{\bm a^\ell}\right)^{\!\top} \pdv{\widetilde{E}}{\bm a^\ell} \leq \left\| \pdv{\mathcal{L}}{\bm a^\ell} \right\|^2~, $
which implies that the gradient of $\mathcal{L}$ is greater than the negative gradient of the residual energy. At the beginning of inference this always holds when using feedforward-pass initialization of hidden activities, since then ${\partial\widetilde{E}}/{\partial\bm a^\ell}=0$. Indeed, \cite{millidge_theoretical_2023} show that this keeps being true also in practice throughout inference, with a sufficiently small learning rate (up to 0.1, which is 2-3 orders of magnitude larger than step sizes typically used in practice). If the learning step is additionally written as a minimization of $\mathcal{L}$, \cite{millidge_theoretical_2023} show that IL will lead to the same minima as BP, even though the weight updates of BP and IL are different. This helps explain the fact that in practice, performance of IL very closely matches BP \cite{whittington_approximation_2017,alonso_theoretical_2022, rosenbaum_relationship_2022,salvatori_stable_2024,song_inferring_2024}. Consequently, deep PCNs could potentially achieve the same generalization performance as BP while maintaining their advantages. 

This result seems to be contradicted in part by \cite{frieder_bad_2024} however, which shows that use of activation functions with many zeros (such as ReLU) can be problematic for PCNs, since they can prevent convergence of learning. This result is interesting in light of \cite{innocenti_understanding_2023} which highlights how ReLU does not show the same speed benefits as other activation functions. Interestingly, even for ANNs, more sophisticated treatments indicate that ReLU is a suboptimal choice of activation function due to the fact that the critical regime {(i.e., the set of initialization parameters at which the network is trainable)} reduces to a single point in phase space \cite{roberts_principles_2022}. In general, these works show how although PCNs and ANNs are similar in many respects, one cannot rely completely on existing knowledge about BP-based frameworks for predicting properties of PCNs.

\section{Conclusion}\label{sec:conclusion}

Despite the remarkable successes of deep learning witnessed in recent years, biological learning is still superior to machine learning (ML) in many key respects, such as energy and data efficiency \cite{zador_catalyzing_2023}. Whereas the most successful deep  learning approaches rely on backpropagation (BP) as a training algorithm, recent work \cite{song_inferring_2024} identified the learning mechanisms in \textit{predictive coding networks} (PCNs), trained with inference learning (IL), to provide an improved model for many features of biological learning.

Research on PCNs has recently witnessed a burst in activity. These networks build on the neuroscientific theory of predictive coding (PC), thus exemplifying recent calls for renewed emphasis on {neuroscience-inspired} methods to artificial intelligence, called \textit{NeuroAI} \cite{zador_catalyzing_2023}. PCNs entail a flexible framework for ML that goes beyond traditional (BP-trained) artificial neural networks (ANNs). Despite the recent burst in activity, a comprehensive and mathematically oriented introduction aimed at machine learning practitioners has been lacking, which this tutorial has attempted to provide.

This work has discussed three complementary perspectives on PCNs (cf. fig. \ref{fig:equilateral_triangle}): 
\begin{enumerate}
    \item a model that generalizes the structure of ANNs (the most recent notion, due to \cite{salvatori_learning_2022});
    \item a probabilistic latent variable model for unsupervised learning (its inception \cite{rao_predictive_1999}); 
    \item a learning algorithm (IL) that can be compared to BP (the most common conception in the recent literature, as for example in \cite{song_inferring_2024}).
\end{enumerate}

The first perspective, PCNs as a generalized ANN, is justified by the fact that mathematically, the structure of PCNs forms a superset of traditional ANNs (a fact that follows from earlier work, but has not been pointed out as a general conception of PCNs). Thus, we expect PCNs share the appealing property of being universal function approximators, while also providing a broader, more general framework as a neural network. Namely, a simple choice of direction gives networks appropriate not only for supervised learning and classification, but also unsupervised learning in generative models. Next to this, the generalization of PCNs to \textit{arbitrary} graphs (PC graphs) \cite{salvatori_learning_2022} means a large new set of structures untrainable by BP can be studied theoretically and empirically. Given the topological nature of several advances in machine learning in the past (such as residual networks \cite{zagoruyko_wide_2016} and the notion of \textit{depth} itself \cite{prince_understanding_2023}), this suggests an interesting avenue for further work.

The second perspective, PCN as probabilistic latent variable model, provides a principled derivation of equations that underlie PCNs. In addition, it brings to light the mentioned connections to other ML algorithms. PCNs share several properties both with classic methods such as factor analysis, but also to modern methods such as diffusion models --  associated with the remarkable success of generative AI witnessed in recent years \cite{prince_understanding_2023}. In general, PCNs' probabilistic nature and these various connections also suggest they could potentially be developed further as a method for \textit{Bayesian deep learning}, which aim to reliably assess uncertainties and incorporate existing knowledge \cite{papamarkou_position_2024, wang_survey_2020}.

The third perspective has been the focus of most of the recent literature on PCNs \cite{whittington_approximation_2017, rosenbaum_relationship_2022, millidge_backpropagation_2023}. Existing implementations of IL are still inefficient compared to BP, which has prevented adoption of PCNs for practical applications, but recent work has taken important steps towards improving efficiency, showing that IL no longer scales with the depth of the network (i.e. it is an order $L$ network depth faster per weight update), if sufficient parallelization is achieved \cite{salvatori_stable_2024}. The reason for this is IL's local computations, which could bring advantages as deep networks keep scaling up, or on neuromorphic hardware. Barring the issue of efficiency, IL demonstrates favorable properties which call for further investigation. In particular, decreased weight interference \cite{song_inferring_2024} and sensitivity to second-order information \cite{alonso_theoretical_2022, innocenti_understanding_2023}, which can lead to faster convergence and increased performance on certain learning tasks (e.g. continual learning and adaptation to changing environments). To what extent these results generalize and scale remains to be studied, but theoretical results suggest that IL should perform at least as well as BP in terms of generalization, potentially opening the door to additional advantages \cite{millidge_theoretical_2023}. 

Finally, we remark that although we have so far refrained from discussing aspects of PC related to biology, psychology, and philosophy, we argue that a ML/computer science perspective is useful for these fields too, since formal detail can help clarify PC's computational properties. This tutorial can be used to review the mathematics used in some of the more speculative works on PC in these domains.

In sum, this tutorial offers a formal overview of PCNs, complementing existing surveys with a detailed mathematical introduction and connections to other ML methods. It clarifies and structures various perspectives, making it an accessible starting point for researchers. As such, it can aid in accelerating future work in the emerging field of NeuroAI.

\begin{acks}
B.v.Z. and E.L.v.d.B. thank our partners, members from the HONDA project between the Honda Research Institute in Japan and Utrecht University in the Netherlands, for supporting this research. The funders had no role in study design, data collection and analysis, decision to publish or preparation of the manuscript. B.v.Z. thanks Lukas Arts and Mathijs Langezaal for helpful discussions and feedback.
\end{acks}

\bibliographystyle{ACM-Reference-Format}
\bibliography{references}

\newpage

\appendix

\renewcommand{\theequation}{A.\arabic{equation}}
\setcounter{equation}{0} 



\section{Expectation Maximization and Variational Inference}\label{sec:app_Bayes}
This appendix extends the discussion in section \ref{sec:probmodel}. We first explain why calculating the likelihood is intractable. Then, we discuss the EM algorithm in mathematical detail. Finally, we discuss variational inference, both for the PCN of Rao \& Ballard, and multi-layer PCNs. We derive the variational free energy for both the case of a delta function and a Gaussian variational distribution.

\subsection{Expectation Maximization}
Given a datapoint $\bm{x}^{(n)}$, we wish to find the maximum likelihood estimate of the parameters $\theta$, defined as those which maximizes the probability of the observed data:
\begin{equation}
\begin{aligned}
\hat{\theta} &= \underset{\theta}{\operatorname{argmax}}\, p(\bm{x}^{(n)}|\theta)\\
	     &= \underset{\theta}{\operatorname{argmin}}\,\underbrace{\left[-\ln p_\theta(\bm{x}^{(n)})\right]}_{\operatorname{NLL}(\theta)}~,
\end{aligned}
\label{eq:MLE}
\end{equation}
where we defined $\operatorname{NLL}(\theta)$ as the negative log-likelihood cf. \eqref{eq:NLL}. Many deep learning models increase their expressivity through \textit{latent} (hidden) variables $\bm z$, low-dimensional compared to $\bm x$. The idea of such variables is that they capture information that is not directly observable in $\bm x$, and they have a simpler probability distribution by design \cite{prince_understanding_2023}. Latent variable models underlie much of the power of modern deep learning methods, in particular generative models \cite{goodfellow_deep_2016, prince_understanding_2023}.

The relation between observed and latent variables is described by the joint distribution $p_\theta(\bm x, \bm z)$. This is also called a \textit{generative model}, because by sampling from it one can generate synthetic data points in the input space $\bm x$. From this one can obtain $p_\theta(\bm x)$ by marginalizing over $\bm z$,
\begin{equation}
	p_\theta(\bm{x}) = \int\!\mathrm{d}\bm z\, p_\theta(\bm x, \bm z)~.
	\label{eq:marginal_likelihood}
\end{equation}
This integral is often intractable \cite{prince_understanding_2023}. Indeed, for PC this is also the case, which can be observed by filling in $p_\theta(\bm x, \bm z)= {\exp}(- E_\theta(\bm x, \bm z))$, and with the energy of Rao \& Ballard, cf. \eqref{eq:RBenergy} one has 
\begin{equation}
\begin{aligned}
p_\theta(\bm x)&= \int\mathrm{d}\bm z p(\bm x, \bm z)\\
&\propto\int \mathrm{d} \bm z \exp \left( \frac{1}{2} \left( \frac{1}{\sigma_x^2}(\bm x - f(W \bm z))^2 + \frac{1}{\sigma_z^2}(\bm z - \bm z^p)^2 \right) \right).
\end{aligned}
\end{equation}
which is generally intractable due to the nonlinearity in the integrand. The same holds for the multi-layer network with \eqref{eq:E_with_C}, where the integral becomes even more complicated. Hence, we cannot just minimize $\text{NLL}(\theta)$ directly. For general generative models, this can be solved in different ways -- commonly through MCMC methods, or variational inference \cite{goodfellow_deep_2016}. \textit{Expectation Maximization} (EM), presented in \eqref{eq:E-step2} and \eqref{eq:M-step2} can be seen as a special case of variational inference \cite{dempster_maximum_1977, bishop_pattern_2006}. Here, we shall motivate EM in detail, highlighting the different steps required to get to IL. 






\subsubsection*{EM in general}

First, introduce some distribution $q_\phi(\bm z)$ over the latent variables. Given a datapoint $\bm{x}^{(n)}$, EM proceeds in two steps \cite{mehta_high-bias_2019, dempster_maximum_1977}. (1) E-step: given $\bm x$ and the current parameters $\theta$, construct a new distribution $q_\phi(\bm z)$ of latent variables, using the current posterior $p_\theta(\bm z|\bm x)$. (2) M-step: re-estimate the parameter $\theta$ to be those with maximum likelihood, assuming $q_\phi(\bm z)$ found in the previous step is the true distribution of $\bm z$. In other words:
\begin{align}
q_\phi(\bm z) &= p_\theta(\bm z|\bm x) && \textbf{(E-step)}, \label{eq:E-step} \\
\hat{\theta} &=\underset{\theta}{\operatorname{argmax}}\, \mathbb{E}_q[\ln p_\theta(\bm x, \bm z)] && \textbf{(M-step)}. \label{eq:M-step}
\end{align}
Taking the $k$-means clustering algorithm as an example, these two steps correspond to assigning points to clusters, and defining clusters based on points, respectively \cite{mackay_information_2003}. This process works because one can show that each EM iteration decreases the true NLL, or at worst leaves it unchanged \cite{dempster_maximum_1977}. This can be shown by writing both steps as minimization of a single functional: the variational free energy.

\subsubsection*{EM with variational free energy}
Define the \textit{variational free energy} $\mathcal{F}$ as follows:
\begin{equation}
	\begin{aligned}
		\mathcal{F}&\coloneqq D_{\text{KL}}[q_\phi(\bm z)||p_\theta(\bm x,\bm z)]\\
		&=D_{\text{KL}}[q_\phi(\bm z)||p_\theta(\bm z|\bm x)] - \ln p_\theta(\bm x)~
	\end{aligned}
\end{equation}
(derived below, cf. section \ref{sec:general_idea}). Since KL-divergences are always positive, one sees that $\mathcal{F} \geq - \ln p_\theta(\bm x)$, i.e., $\mathcal{F}$ is an \textit{upper bound} on the NLL. In other words, if minimized with respect to $\theta$, the NLL is also minimized. Similarly, it can be shown that the E-step can be written as minimization of this function \cite{neal_view_1998}. 
Thus, one can write a new version of EM as follows:
\begin{align}
q_\phi(\bm z) &= \underset{q}{\operatorname{argmin}}\, \mathcal{F}(q,\theta) && \textbf{(E-step)}, \\ 
\hat{\theta}&=\underset{\theta}{\operatorname{argmin}}\, \mathcal{F}(q,\theta) && \textbf{(M-step)}.
\end{align}
Below we demonstrate how different choices of $q$ lead to different VFEs. For now, it can be taken simply as an intuitive justification for why the EM process minimizes the NLL as desired.

\subsubsection*{EM with MAP estimation}
Using the result in \cite{neal_view_1998} that EM will minimize the NLL, we return to the first formulation, \eqref{eq:E-step} and \eqref{eq:M-step}. At first sight, the E-step still looks problematic, since evaluating the posterior $p_\theta(\bm z| \bm x)=p_\theta(\bm x, \bm z)/p_\theta(\bm x)$ requires calculating the marginal likelihood $p_\theta(\bm x)$. This is analogous to computing the partition function in statistical physics, which is generally intractable. A solution 
is to set $q$ to a delta function, $q_\phi(\bm z) = \delta(\bm z-\widetilde{\bm z})$, in which case the posterior is simply described by its maximum value. Put differently, we simplify the evaluation of the posterior in the E-step by a \emph{maximum a posteriori (MAP)} estimation of $p(\bm z|\bm x)$. With a delta-distribution, the expectation of \eqref{eq:M-step} also simplifies, and we have
\begin{align}
    \widetilde{\bm z} &= \underset{\bm z}{\operatorname{argmax}}\,p_\theta(\bm x,\bm z)&& \textbf{(E-step)},    \\
\hat{\theta}&=\underset{\theta}{\operatorname{argmax}}\, p_\theta(\bm x, \widetilde{\bm z}) && \textbf{(M-step)}.
\end{align}
With this choice, the parameters of $q_\phi(\bm z)$ are $\phi=\widetilde{\bm z}$. We discuss this in more detail below.
Thus, we see having a model for $\ln p_\theta(\bm x, \bm z)$ is sufficient for minimizing the (upper bound of) the NLL  using EM. 
By choosing a form for $p_\theta(\bm x, \bm z)$ (or $E$) one obtains different models, including PC.

\subsubsection*{EM with partial steps}
Writing EM as a minimization of the VFE, it can be shown that minimizations in both steps need not be \textit{complete}, but can also be \textit{partial} \cite{neal_view_1998}. 
Taking steps that decreases $\mathcal{F}$ (i.e., in a gradient descent formulation) will also minimize the NLL. Algorithms with partial E and/or M steps are called \textit{generalized EM} algorithms \cite{neal_view_1998, bishop_pattern_2006}. Thus, we can write our process using a partial M-step:
\begin{align}
\widetilde{\bm z} &= \underset{\bm z}{\operatorname{argmin}}\,E(\bm x, \bm z)&& \textbf{(E-step)} \\
\Delta {\theta} &\propto -\frac{dE}{d{\theta}}&& \textbf{(partial M-step)} 
\end{align}
this is standard IL as presented in section \ref{sec:ANN_to_PCN}. Alternatively, we could also make the E-step partial:
\begin{align}
\Delta \bm{z} &\propto  -\frac{dE}{d\bm{z}}&& \textbf{(partial E-step)},    \\
\Delta {\theta} &\propto -\frac{dE}{d {\theta}}&& \textbf{(partial M-step)},
\end{align}
which is the basis for incremental IL.


\subsection{Variational Inference}\label{sec:VI}
Variational inference provides a more general understanding for why EM minimizes the NLL. It also provides a connection to the more general theory of variational methods used in statistical physics and machine learning, for instance in variational autoencoders (VAEs). We briefly explain the general idea.

\subsubsection{General Idea}\label{sec:general_idea}
Consider Bayes' rule for the posterior $p(\bm z|\bm x)$:
\begin{equation}p_\theta(\bm z|\bm x)=\frac{p_\theta(\bm x,\bm z)}{p_\theta(\bm x)}. 
\label{eq:Bayes}
\end{equation}
Ideally, we would like to compute this posterior, but this requires computing the prior distribution $p_\theta(x)$, which is typically intractable. Instead, pariational methods attempt to approximate the posterior distribution with a so-called \textit{variational posterior }$q_\phi(\bm z)$ parametrized by $\phi$, such that the KL-divergence between $q_\phi$ and $p(\bm z|\bm x)$,
\begin{align*}
D_{\text{KL}}[q_\phi(\bm z)||p_\theta(\bm z|\bm x)] &=\int\!\mathrm{d}\bm z \,q_\phi(\bm z) \ln \frac{q_\phi(\bm z)}{p_\theta(\bm z|\bm x)}
\label{eq:VFE}
\end{align*}
is minimized. By Bayes' rule, this becomes
\begin{align*}
D_{\text{KL}}[q_\phi(\bm z)||p_\theta(\bm z|\bm x)] &=\int\!\mathrm{d}\bm z\,q(\bm z) \ln \frac{q_\phi(\bm z)}{p_\theta(\bm x,\bm z)}+ \ln p_\theta(\bm x)\\
&=\underbrace{D[q_\phi(\bm z)||p_\theta(\bm x,\bm z)]}_{=\mathcal{F}}+\ln p_\theta(\bm x)~,
\end{align*}
where we have recognized variational free energy $\mathcal{F}$ introduced above, which we can calculate for a given generative model and variational posterior. Furthermore, since the KL divergence is non-negative, 
\begin{equation}
	\mathcal{F} \geq D[q_\phi(\bm z)||p_\theta(\bm z|\bm x)]~,
	\label{eq:UB1}
\end{equation}
since $\ln p_\theta(x)\leq 0$. Thus, if we minimize $\mathcal{F}$ with respect to $\phi$, our function $q$ will get closer to the true posterior: we are performing variational inference. In addition, we see that $\mathcal{F}$ is a tractable upper bound to the NLL: minimizing $\mathcal{F}$ thus minimizes the NLL. 

In calculations, it is useful to decompose $\mathcal{F}$ into two terms:
\begin{equation}
\mathcal{F}= \underbrace{\int\!\mathrm{d} \bm{z}\,q(\bm z) \ln q(\bm z)}_{H[q(\bm z)]=H[q]}+ \underbrace{\int\!\mathrm{d} \bm{z} \,q(\bm z ) E(\bm x,\bm z ))}_{E_q}
\label{eq:vfe_2terms}
\end{equation}
where we have substituted $E(\bm x,\bm z ) = -\ln p_\theta(\bm x,\bm z )$. To make notation less cluttered, we leave out $\phi$ from here onwards. In this expression one can fill in a generative model and variational posterior, which we do below. First, we consider the general form of the VFE for $L$ latent vectors, which is:
\begin{align*}
\mathcal{F} &= D_{\text{KL}}[q(\{\bm z^\ell \})||p_\theta(\bm{x},\{\bm z^\ell \})].\\
&\equiv \int \prod_{\ell=1}^L d \bm{z}^\ell q(\{\bm z^\ell \})\ln \frac{q(\{\bm z^\ell \})}{p_\theta(\bm x,\{\bm z^\ell \})}\\
&=\underbrace{\int \prod_{\ell=1}^L \mathrm{d}\bm{z}^\ell q(\bar{\bm z}) \ln q(\bar{\bm z})}_{H[\bar{q}]}
	-\underbrace{\int \prod_{\ell=1}^L \mathrm{d} \bm{z}^\ell q(\bar{\bm z}) E(\bm x,\bar{\bm z})}_{E_{\bar{q}}}~,
\end{align*}
where we used $E(\bm{x},\{\bm z^\ell \}) = -\ln p_\theta(\bm{x},\{\bm z^\ell \})$ and introduced the notation $\bar{\bm z} = \{ \bm z^\ell \}$ to make notation less cluttered. 
For the multi-layer case, it is often assumed $q$ factorizes over the layers, i.e.,\footnote{This does not correspond to any realistic network architecture, since it effectively assumes that $q(\bm z^\ell|\bm z^{\ell-1})=q(\bm z^\ell)$, i.e., that the individual layers are completely independent. As a variational ansatz however, it is not unreasonable, at least in certain cases. In FNNs at large width for example, the central limit theorem implies that the layers are approximately Gaussian, so a Gaussian ansatz for each separate layer can still lead to good results.}
\begin{equation}
	\bar{q}= q(\bar{\bm z}) = \prod_{\ell=1}^L q(\bm z^{\ell})~,
 \label{eq:q_factorization}
\end{equation}
whereupon the entropy becomes
\begin{equation}
\begin{aligned}
	H[\bar{q}] &= \int\!\prod_{\ell=1}^{L}\mathrm{d} \bm z^\ell q(\bm z^\ell) \sum_{k=1}^L \ln q(\bm z^k)\\
 &=\sum_k \int\!\mathrm{d} \bm z^k q (\bm z^k ) \ln q (\bm z^k) \underbrace{\prod_{\ell \neq k} \int\!\mathrm{d} \bm z^\ell q (\bm z^\ell)}_{=1}\\
 &= \sum_\ell H[q(\bm z^\ell)]~,
\label{eq:H_qk}
\end{aligned}
\end{equation}
i.e., the entropy factorizes into a sum of the entropies of the individual layers. The second term $E_{\bar{q}}$ is somewhat more complicated in general, since the latent vectors in different layers $\bm z^\ell$ may be coupled through $E$ even though $q(\bar{\bm z})$ factorizes. If one assumes a hierarchical structure of the generative model, cf. \eqref{eq:markov}, one may write:
\begin{equation}
    E = \sum_{\ell=1}^L E(\bm z^\ell|\bm z^{\ell+1})
\end{equation}
where if $E$ is given by \eqref{eq:E_with_C} then $E(\bm z^\ell|\bm z^{\ell+1})=\frac{1}{2}(\bm \epsilon^\ell)^{T} (\Sigma^{\ell})^{-1}\bm \epsilon^\ell + \ln \det \Sigma^\ell$. This provides some simplification in $E_{\bar q}$:
\begin{equation}
\begin{aligned}
E_{\bar q}&= \int \prod_{\ell} \mathrm{d}\bm{z}^\ell q(\bm z^\ell) \sum_{k}E(\bm z^{k}|\bm z^{k+1} )\\
&= \int \mathrm{d} \bm z^1 q(\bm z ^1) E(\bm z^0| \bm z^1)\prod_{\ell\neq1} \underbrace{\int \mathrm{d} \bm z^\ell q(\bm z^\ell)}_{=1} \\
&\quad+\sum_k \int \mathrm{d} \bm z^k \mathrm{d}\bm z^{k+1} q(\bm z ^k) q(\bm z ^{k+1}) E(\bm z^k| \bm z^{k+1})\prod_{\ell \neq k,\,\ell\neq k+1}\underbrace{\int \mathrm{d} \bm z^\ell q(\bm z^\ell)}_{=1}\\
&\quad +\int \mathrm{d} \bm z^L q(\bm z ^L) E(\bm z^L|\bm z^p) \prod_{\ell\neq L} \underbrace{\int \mathrm{d} \bm z^\ell q(\bm z^\ell)}_{=1}\\
&=\underbrace{\int\mathrm{d} \bm z^1 q(\bm z ^1) E(\bm z^0| \bm z^1)}_{(1)}+\sum_\ell \underbrace{\int \mathrm{d}\bm z^\ell d\bm z^{\ell+1} q(\bm z ^\ell) q(\bm z ^{\ell+1}) E(\bm z^\ell| \bm z^{\ell+1})}_{(2)} +\underbrace{\int \mathrm{d} \bm z^L q(\bm z ^L) E(\bm z^L|\bm z^p) }_{(3)}~.
\end{aligned}
\end{equation}
For $L$ latent vectors assuming a hierarchical generative model and a factorized variational, this is the furthest one can simplify. Observe that (1) is identical to $E_q$ in eq. \eqref{eq:vfe_2terms}, whereas (3) is slightly different, and (2) is more complex.


\subsubsection{PCNs of Rao \& Ballard: Delta-function variational} \label{sec:single_layer_VI_delta}

Recent works \cite{millidge_predictive_2022, pinchetti_predictive_2022} present an attempted derivation of PC using a delta function as the variational posterior, in which it is claimed that the VFE simply becomes equal to the energy. Unfortunately however, this derivation contains a basic mathematical error, which we explain below. An alternative derivation \cite{buckley_free_2017} also exists that instead uses a Gaussian as variational posterior, approximated using a Taylor series. Under certain assumptions, this leads to the claimed result. We review the both here and in the next subsection.

First, let us take an $n_z$-dimensional delta function as our variational posterior:
\begin{equation}
	\delta^{(n_z)}(\bm z - \widetilde{\bm z})=\prod_{i=1}^{n_z} \delta (z_i-\widetilde{z}_i)~.
\end{equation}
For the differential entropy $H_q$ in \eqref{eq:vfe_2terms}, this gives
\begin{equation}
	\begin{aligned}
H_q &= \int\!\mathrm{d} \bm z\,\delta^{(n_z)}(\bm z-\widetilde{\bm z}) \ln \delta^{(n_z)}(\bm z-\widetilde{\bm z})\\
&=\int\!\prod_{i=1}^M \mathrm{d}z_i\,\delta(z_i-\widetilde{z}_i) \sum_{j=1}^M\ln \delta(z_j-\widetilde{z}_j)\\
&=\sum_{j=1}^M \biggl( \int\!\mathrm{d}z_j \delta(z_j-\widetilde{z}_j) \ln \delta(z_j-\widetilde{z}_j) \prod_{i \neq j} \underbrace{\int\!\mathrm{d}z_i\,\delta(z_i-\widetilde{z}_i)}_{=1} \biggr)~.
\label{eq:zero}
	\end{aligned}
\end{equation}
In \cite{millidge_predictive_2022} this is erroneously taken to be zero.
We speculate that this is based on the intuition that the ``entropy'' of a deterministic function like $\delta(z)$ should vanish. In fact however, the differential entropy of the Dirac delta function is $-\infty$.
If however one treats this as an additive constant to be dropped, then we have only the $E_q$ term, so that
\begin{equation}
	\mathcal{F} = E(\bm x, \widetilde{\bm z})~.
\end{equation}
For this reason, \cite{millidge_predictive_2022} identify the VFE as simply the energy (or negative complete data log-likelihood) in this case. 

\subsubsection{PCNs of Rao \& Ballard: Gaussian variational} \label{sec:rao_ballard_gaussian}

Now we instead take $q$ to be a multivariate Gaussian centered around $\widetilde{\bm z}$ with covariance $\widetilde \Sigma$, using a tilde to distinguish from the mean and covariance of the generative model. This generalizes the derivation in \cite{buckley_free_2017} to the multivariate case, i.e., 
\begin{equation}
	q(\bm z) = \frac{1}{\sqrt{(2 \pi)^{n_z}\operatorname{det}\widetilde{\Sigma} }}\exp \left(-\frac{1}{2}(\bm{z}-\widetilde{\bm z} )^{T} \widetilde{\Sigma}^{-1}(\bm{z}-\widetilde{\bm z} )\right)~.
\end{equation}
For convenience, we label $a_i = z_i-\widetilde{z}_i$, $A=\widetilde{\Sigma}^{-1}$, and $Z=\sqrt{(2 \pi)^{n_z}\operatorname{det}A^{-1} }$. 
Substituting this ansatz into \eqref{eq:vfe_2terms}, we have
\begin{equation}
	\begin{aligned}
		H[q] &= \int\!\mathrm{d} \bm{z}\,q(\bm z) \ln q(\bm z)\\
		&= -\frac{1}{2Z}\int\!\mathrm{d}\bm a \left[\exp \bigg( -\frac{1}{2} \sum_{ij} a_i a_j A_{ij} \bigg)\sum_{nm} a_n a_m A_{nm}\right]-\ln Z\!\underbrace{\int\!\mathrm{d}\bm{z}\, q(\bm z)}_{=1} \\
		&= -\frac{1}{2Z}\sum_{nm} A_{nm} \underbrace{\int\!\mathrm{d}\bm a \left[ a_n a_m \exp \bigg( -\frac{1}{2} \sum_{ij} a_i a_j A_{ij} \bigg) \right]}_{\langle a_n a_m\rangle}-\ln Z~,
	\end{aligned}
\end{equation}
where we have identified the simple Gaussian integral
\begin{equation}
	\langle a_n a_m\rangle= \sqrt{\frac{(2\pi)^{n_z}}{\det A}} (A^{-1})_{mn}~.
\end{equation}
Plugging this in and writing out the definition of $Z$, we thus obtain
\begin{equation}
	\begin{aligned}
		H[q] &= -\frac{1}{2}\sum_{n}\delta_{nn}-\frac{1}{2}\left[ n_z\ln(2 \pi)+ \ln \operatorname{det}\widetilde{\Sigma}\right]\\
	     	& = -\frac{1}{2} \left[n_z(\ln 2\pi +1)+ \ln \det \widetilde{\Sigma}\right]~.
	\end{aligned}
	\label{eq:H_qz}
\end{equation}
 Meanwhile for $E_q$, one obtains:
 \begin{equation}
\begin{aligned}
E_q & = \int\!\mathrm{d} \bm{z}\,q(\bm z ) E(\bm x,\bm z ) \\
&=\frac{1}{2Z}\int\!\mathrm{d}\bm z \exp \left(-\frac{1}{2}(\bm z - \widetilde{\bm z})^T\widetilde{\Sigma}^{-1}(\bm z - \widetilde{\bm z})\right)\bigg[(\bm x - f(W\bm z))^T\Sigma_x(\bm x - f(W\bm z))\\
&\quad \quad+ (\bm z - \bm z^p)\Sigma_z(\bm z - \bm z^p)+ D \bigg]~,
\end{aligned}
 \end{equation}
 where $D= \frac{1}{2}\left((n_x+n_z)\ln 2\pi+ \ln \left[\det \Sigma_x \det \Sigma_z\right]\right)$. This integral does not have a full closed-form solution due to the nonlinearity $f(W\bm z)$. In the absence of a specific expression for this function, there are several ways to proceed. If $f(W\bm z)$ admits a polynomial expansion in $\bm z$, then all odd terms in the integrand vanish, and the resulting integrals of the form $\int\!\mathrm{d}\bm z\,{\bm z}^n\,q(\bm z)$ for $0<n\in2\mathbb{Z}_+$ are easily done. A more abstract way of proceeding would be to perform an Edgeworth expansion of $p(\bm x,\bm z)$, with some assumption about how the corresponding cumulants integrate against a Gaussian. Intermediate between these two is the approach taken by \cite{buckley_free_2017}, which keeps $E$ unspecified, and formally Taylor expands it to second order around the mean of the variational posterior $\bm z = \widetilde{\bm z}$. However, this implicitly assumes that $f(W\bm z)$ does not grow too fast in $\bm z$ (i.e., the polynomial expansion quickly truncates); otherwise, the fall-off of the Gaussian $q(\bm z)$ is insufficiently fast for this approximation to apply for general functions. Furthermore, even in the special case where $\langle E(\bm x,\bm z)\rangle_z= \widetilde{\bm z}$, there is no guarantee that expanding the energy around this point will yield a good approximation to the integral. If $E(\bm x,\bm z)$ is sharply peaked at $\widetilde{\bm z}$ however, then we may proceed as in \cite{buckley_free_2017}; to second order, 
\begin{equation}
\begin{aligned}
	E(\bm x, \bm{z})&\approx E(\bm x, \widetilde{\bm z}) + \nabla_{\bm z}E(\bm x, \widetilde{\bm z}) + \frac{1}{2}(\bm z- \widetilde{\bm z})^T \nabla\nabla_{\bm z} E(\bm x, \widetilde{\bm z})(\bm z- \widetilde{\bm z})\\
&= E(\bm x, \widetilde{\bm z}) + \sum_i \frac{\partial E(\bm x, \widetilde{\bm z})}{\partial z_i} (z_i-\widetilde{z}_i)+\frac{1}{2}\sum_{ij} \frac{\partial^2 E(\bm x, \widetilde{\bm z})}{\partial z_i \partial z_j} (z_i-\widetilde{z}_i) (z_j-\widetilde{z}_j)~.
\end{aligned}
\end{equation}
We define $H =\nabla\nabla_{\bm z}E(\bm x, \widetilde{\bm z})$ for the Hessian (the matrix of second order derivatives). This gives
\begin{equation}
\begin{aligned}
	E_q =&\int\!\mathrm{d} \bm{z} \,q(\bm{z} ) E(\bm{x},\bm{z} ) \\
	\approx&\,E(\bm x, \widetilde{\bm z}) \underbrace{\int\!\mathrm{d} \bm{z}\,q(\bm z )}_{=1} + \sum_i \frac{\partial E(\bm x, \widetilde{\bm z})}{\partial z_i} \underbrace{\int\!\mathrm{d}\bm{z}\,q(\bm z ) (z_i-\widetilde{z}_i)}_{=0}\\
	    &+\frac{1}{2}\sum_{ij} \frac{\partial^2 E(\bm x, \widetilde{\bm z})}{\partial z_i \partial z_j} {\int\!\mathrm{d} \bm{z}\,q(\bm z )(z_i-\widetilde{z}_i) (z_j-\widetilde{z}_j)} ~,
\end{aligned}
\end{equation}
where the second term vanishes since the integrand is odd. Defining the shorthand notation
$z_j-\widetilde{z}_j \eqqcolon a_j$, the remaining integral is 
\begin{equation}
\begin{aligned}
\int\!\mathrm{d} \bm{z}\,q(\bm z )(z_i-\widetilde{z}_i) (z_j-\widetilde{z}_j) &= Z^{-1}\int\!\mathrm{d}\bm a \exp \bigg(-\frac{1}{2} \sum_{kl}a_k a_l A_{kl}\bigg)a_ia_j \\ 
&=Z^{-1} \sqrt{\frac{(2\pi)^{n_z}}{\det A}} (A^{-1})_{ij}
=(A^{-1})_{ij}=\widetilde{\Sigma}_{ij}~,
\end{aligned}	
\label{eq:star}
\end{equation}
and therefore
\begin{equation}
E_q \approx E(\bm x, \widetilde{\bm z}) + \frac{1}{2}\sum_{ij} H_{ij}\widetilde{\Sigma}_{ij}~.
\label{eq:Eq_1layer}
\end{equation}
Collecting results, the variational free energy $\mathcal{F}$ is then
\begin{equation}
\mathcal{F} = E(\bm x, \widetilde{\bm z}) + \frac{1}{2}\sum_{ij} H_{ij} \widetilde{\Sigma}_{ij} -\frac{1}{2} \ln \det \widetilde{\Sigma} -\frac{1}{2} n_z(\ln 2\pi +1)~.
\end{equation}
Thus, in contrast to the approach via a delta function above, a Gaussian ansatz leads to a term for the covariance of the variational posterior, as well as higher-order derivatives of the energy function. Applying EM would then require an additional part of the E-step for $\widetilde{\Sigma}$. However, minimizing $\mathcal{F}$ with respect to $\widetilde{\Sigma}$ yields
\begin{equation}
	\pdv{\mathcal{F}}{\Sigma_{ij}} = \frac{1}{2}H_{ij}-\frac{1}{2}(\widetilde{\Sigma}^{-1})_{ij}~,
\end{equation}
where we used the identity $\pdv{\ln \det A}{A_{ij}}= (A^{-1})_{ji}$, which follows from Jacobi's formula, and symmetry of the covariance matrix. Thus, $\mathcal{F}$ is optimized when: 
\begin{equation}
	\widetilde\Sigma_{ij} = (H^{-1})_{ij}~,
\end{equation}
for all $i,j$. This gives 
\begin{equation}
\begin{aligned}
    \mathcal{F} &= E(\bm x, \widetilde{\bm z}) + \frac{1}{2}\underbrace{\sum_{ij} H_{ij} (H^{-1})_{ij}}_{=n_z} -\frac{1}{2} \ln \det H^{-1} -\frac{1}{2} n_z(\ln 2\pi +1)~\\
    &=E(\bm x, \widetilde{\bm z}) -\frac{1}{2} \ln \det H^{-1} -\frac{1}{2} n_z\ln 2\pi~
\end{aligned}
\end{equation}
where we used the symmetry of the Hessian. This is the multivariate counterpart to eq. (17) in \cite{buckley_free_2017}. In sum, under a second order Taylor approximation of $E$, the optimal covariance of the variational is equal to the Hessian of $E$. Interestingly, including this term was shown to have practical consequences in \cite{zahid_curvature-sensitive_2023} training generative PCNs, with substantially improved sample qualities and log-likelihoods as a result. Hence, further studying the inclusion of higher-order terms in the minimization objective could potentially be an interesting avenue for further work. However, as noted by \cite{zahid_curvature-sensitive_2023}, such efforts need to balance potential increases in performance with increased computation. Moreover, better variational ans\"atze may encounter other practical issues, cf. \cite{mackay_information_2003, mackay_problem_2001}.

\section{Complementary Proofs and Miscellaneous}
\subsection{Backpropagation}\label{sec:app_BP}
We derive \eqref{eq:BP_error1-5} and \eqref{eq:BP_errpr} from \eqref{eq:activity_FNN} in section \ref{sec:discriminative_PCNs}. Observe that with the chain rule, \eqref{eq:BP_errpr} can be written as:
\begin{equation}
\begin{aligned}
  {\delta}^\ell_i &= \sum_j\frac{\partial\mathcal{L}}{\partial {a}^{\ell+1}_j}\frac{\partial a_j^{\ell+1}}{\partial {a}_i^{\ell}}\\
&= \sum_j \delta_j^{\ell+1}f'\Big(\sum_k {w}_{jk}^{\ell} {a}_k^{\ell}\Big) w_{ji}^{\ell}  
\end{aligned}
\end{equation}
I.e. the error in layer $\ell$ can be expressed recursively using the error in the next layer, $\ell+1$. Then, observe that the derivative with respect to weights can be expressed using this error:
\begin{equation}
    	\frac{\partial\mathcal{L}}{\partial {w}_{ij}^\ell} = \sum_k \frac{\partial\mathcal{L}}{\partial {a}_k^{\ell+1}} \frac{\partial {a}_k^{\ell+1}}{\partial {w}_{ij}^\ell} 
    	=  \delta_i^{\ell+1}f'\Big(\sum_n {w}^{\ell}_{in} {a}_n^{\ell} \Big)a_j^\ell~,
\end{equation}
which using matrix notation gives the results in the main text.

\subsection{Equivalence of FNNs and Discriminative PCNs During Testing}\label{sec:app_proof_FNN_PCN}
This appendix reproduces the proof in \cite{whittington_approximation_2017,song_predictive_2021} that inference (testing) in FNNs and discriminative PCNs is equivalent, cf. section \ref{sec:testing_PCNs} in the main text. During testing, clamp $\bm{a}^0 = \bm{x}^{(n)}$. We want to find
$$\hat{\bm{a}}^\ell = \underset{\bm{a}^\ell}{\operatorname{argmin}} \, {E}(\bm{a}, \bm{w})$$
for $\ell=1,...,L$. The inferred value of the last layer $\bm{a}^L$ is identified as the output $\hat{\bm{y}}$. Write the energy as:
\begin{equation}
    E(\bm{a}, \bm{w}) = \frac{1}{2} \sum_{\ell=1}^L (\bm{\epsilon}^\ell)^2 = \frac{1}{2} \left( (\bm{\epsilon}^{L})^2 + (\bm{\epsilon}^{L-1})^2 +\sum^{L-2}_{\ell=1} (\bm{\epsilon}^\ell)^2 \right) 
\end{equation}
and note that $\bm{\epsilon}^\ell=\bm{a}^{\ell}-f(\bm{w}^{\ell-1}\bm{a}^{\ell-1})$. To find the minimum, take the derivative w.r.t. $\bm{a}^\ell$ and set to zero, for all $\ell=1,...,L$. First for $\ell=L$:
\begin{equation}
    \frac{\partial E}{\partial\bm{a}^L} = \bm{\epsilon}^L = \bm{a}^{L}-f(\bm{w}^{L-1}\bm{a}^{L-1}) = 0 \iff \bm{a}^{L}=f(\bm{w}^{L-1}\bm{a}^{L-1})~.
    \label{eq:dEdaL}
\end{equation}
For $\ell=L-1$:
\begin{equation}
    \frac{\partial E}{\partial\bm{a}^{L-1}} = \bm{\epsilon}^{L}\frac{\partial \bm{\epsilon}^{L}}{\partial\bm{a}^{L-1}} + \bm{\epsilon}^{L-1}=0~.
\end{equation}
Now from \eqref{eq:dEdaL} we know that $\bm{\epsilon}^L=0$, and hence we have 
\begin{equation}
    \bm{\epsilon}^{L-1}=0 \iff \bm{a}^{L-1}=f(\bm{w}^{L-2}\bm{a}^{L-2}).
\end{equation}
Continuing this argument for decreasing $\ell$ until $\ell=0$ we obtain
\begin{equation}
    \bm{\epsilon}^{1}=0 \iff \bm{a}^{1}=f(\bm{w}^{0}\bm{a}^{0})~.
\end{equation}
Thus we have found that for $\ell=L,...,1$,
\begin{equation}
    \bm{a}^{\ell}=f(\bm{w}^{\ell-1}\bm{a}^{\ell-1}), 
\end{equation}
so with $\bm{a}^{0}=\bm{x}^{(n)}$ and $\bm{a}^{L}=\hat{\bm{y}}$ we obtain exactly the relations for a FNN, cf. \eqref{eq:activity_FNN}. 
\subsection{Comparison Tables \& Algorithms}\label{sec:comparisons}
Tables \ref{tab:comp_FNN_PCN} and \ref{tab:comp_BP_IL} show side-by-side comparisons of the computations involved in training and testing a FNN/PCN, as well as a more detailed comparison of BP and IL. Table \ref{tab:PCG_Vs_PCN} compares hierarchical PCNs to PCGs.

\begin{table} 
\centering
\caption{\textbf{Training \& testing procedures compared}: FNNs vs. discriminative PCNs. Key steps and corresponding equations are shown, and biases are omitted for clarity. {Observe the difference between a prediction $\bm \mu^\ell=f(\bm w^{\ell-1}\bm a^{\ell-1})$ in PC, and the activity rule in an FNN,  $\bm a^\ell=f(\bm w^{\ell-1}\bm a^{\ell-1})$.} `{Fixed} output' refers to the final activity layer $\bm{a}^L$, clamped to data $\bm{y}^{(n)}$ during training (absent in FNNs). `{Predicted} output' is the prediction in the final layer, $\bm \mu^L$, which becomes an output during testing, $\bm \mu^L=\hat{\bm{y}}$.}
\setcellgapes{6pt}
\makegapedcells
\resizebox{\columnwidth}{!}{%
\begin{tabular}{lllll}
 &
  \multicolumn{2}{c}{\textit{Training Procedure}} &
  \multicolumn{2}{c}{\textit{Testing Procedure}} \\  
  \hline
\multicolumn{1}{l}{} &
  \textbf{FNN} &
  \multicolumn{1}{l}{\textbf{Discriminative PCN}} &
  \textbf{FNN} &
  \multicolumn{1}{l}{\textbf{Discriminative PCN}} \\ \hline \hline
\multicolumn{1}{|l|}{Input (fixed)} &
  $\bm{a}^0=\bm{x}^{(n)}$ &
  \multicolumn{1}{l|}{$\bm{a}^0=\bm{x}^{(n)}$} &
  $\bm{a}^0=\bm{x}^{(n)}$ &
  \multicolumn{1}{l|}{$\bm{a}^0=\bm{x}^{(n)}$} \\ \hline
\multicolumn{1}{|l|}{Output (fixed)} &
  $-$ &
  \multicolumn{1}{l|}{$\bm{a}^L=\bm{y}^{(n)}$} &
  $-$ &
  \multicolumn{1}{l|}{$-$} \\ \hline \hline
\multicolumn{1}{|l|}{Activity rule} &
  $\bm{a}^\ell=f(\bm{w}^{\ell-1}\bm{a}^{\ell-1})$ &
  \multicolumn{1}{l|}{$\hat{\bm{a}}=\underset{\bm{a}}{\operatorname{argmin}} \,   E(\bm{a},\bm{w})$} &
  $\bm{a}^\ell=f(\bm{w}^{\ell-1}\hat{\bm{a}}^{\ell-1})$ &
  \multicolumn{1}{l|}{$\begin{aligned}
  \hat{\bm{a}}&=\underset{\bm{a}}{\operatorname{argmin}}\,E(\bm{a},\bm{w})\\&=f(\bm{w}^{\ell-1}\hat{\bm{a}}^{\ell-1})\end{aligned}$} \\ \hline
\multicolumn{1}{|l|}{Output (predicted)} &
  $\hat{\bm{y}}=\bm{a}^L$ &
  \multicolumn{1}{l|}{$\hat{\bm{y}}=\bm{\mu}^L$} &
  $\hat{\bm{y}}=\bm{a}^L=f(\bm{w}^{L-1}\bm{a}^{L-1})$ &
  \multicolumn{1}{l|}{$\hat{\bm{y}}=\bm{\mu}^L=f(\bm{w}^{L-1}\hat{\bm{a}}^{L-1})$} \\ \hline \hline
\multicolumn{1}{|l|}{Objective} &
  $(\hat{\bm{y}}-\bm{y}^{(n)})^2$ &
  \multicolumn{1}{l|}{$(\hat{\bm{y}}-\bm{y}^{(n)})^2+\sum\limits_{\ell=1}^{L-1}(\bm{\epsilon}^\ell)^2$} &
  $(\hat{\bm{y}}-\bm{y}^{(n)})^2$ &
  \multicolumn{1}{l|}{$(\hat{\bm{y}}-\bm{y}^{(n)})^2$} \\ \hline
\multicolumn{1}{|l|}{Learning rule} &
  $\hat{\bm{w}}^\ell = \underset{\bm{w}^\ell}{\operatorname{argmin}} \,   \mathcal{L}(\bm{w})$ &
  \multicolumn{1}{l|}{$\hat{\bm{w}}^\ell = \underset{\bm{w}^\ell}{\operatorname{argmin}}   \,   E(\hat{\bm{a}},\bm{w})$} &
  $-$ &
  \multicolumn{1}{l|}{$-$} \\ \hline \hline
\end{tabular}
}
\vspace{0.1cm}

\label{tab:comp_FNN_PCN}
\end{table}

\begin{table}[] 
\centering
\caption{\textbf{Learning algorithms compared}: backprop (BP) in an FNN vs. inference learning (IL) in a discriminative PCN. Whereas the form of equations are similar in both algorithms, they are conceptually different. BP proceeds in a backward and forward phase, such that nodes in lower layers need to wait for the error computed at the output to be propagated back. IL has no such waiting times since layers can be updated in parallel. At the same time, processing a batch in BP is done by a single forward and backward pass, while in IL this requires several inference. {Note that the form of the equations changes depending on the convention chosen, cf. section \ref{sec:direction}.}}
\setcellgapes{6pt}
\makegapedcells
\resizebox{\columnwidth}{!}{%
\begin{tabular}{|ll|ll|}
\multicolumn{2}{c}{\textbf{Backprop}}                                                                                                                                                                             & \multicolumn{2}{c}{\textbf{Inference Learning}}                                                                                                                  \\ 
\hline \hline
Forward pass  & $\bm{a}^\ell=f(\bm{w}^{\ell-1}\bm{a}^{\ell-1}) \quad\quad \ell:1\rightarrow L$                                                                                                                                                   & Inference & $\bm{\mu}^\ell=f(\bm{w}^{\ell-1}\bm{a}^{\ell-1})$                                                                                                    \\ 
\cline{1-2}
Backward pass & \multirow{2}{*}{$\begin{rcases*}
\bm{\delta}^\ell&$=\frac{\partial\mathcal{L}}{\partial z^\ell}$\\
&$=(\bm{w}^\ell)^T\bm{\delta}^{\ell+1}\odot f'(\bm{w}^\ell\bm{a}^\ell)$
\end{rcases*}\, \ell: L\rightarrow 1$} &           & $\bm{\epsilon}^\ell=\bm{a}^\ell-\bm{\mu}^\ell$                                                                                                       \\
              &                                                                                                                                                                                                   &           & $\Delta \bm{a}^\ell= -\gamma\left( \bm{\epsilon}^\ell-(\bm{w}^\ell)^T \bm{\epsilon}^{\ell+ 1}\odot f'\left( \bm{w}^\ell\bm{a}^\ell   \right) \right)$  \\ 
\hline
Learning      & $ \Delta \bm{w}^\ell=\alpha \bm{\delta}^{\ell+1}\odot f'(\bm{w}^\ell\bm{a}^\ell)(\bm a^\ell)^T$
	      & Learning  & $\Delta \bm{w}^\ell = \alpha \bm{\epsilon}^{l+1}\odot f'\left(\bm{w}^\ell\hat{\bm{a}}^\ell\right)(\hat{\bm{a}}^\ell)^T$                              \\
\hline \hline
\end{tabular}
}

\label{tab:comp_BP_IL}
\end{table}

\begin{table}[]
\centering
\caption{\textbf{Hierarchical PCNs compared to PC graphs.} The latter is a superset of the former insofar as it allows arbitrary connectivity between nodes.}
\setcellgapes{4pt}
\makegapedcells
\begin{tabular}{lll}
\hline
           & \textbf{Hierarchical PCN }                                      & \textbf{PC Graph}                            \\ \hline
Prediction & $\bm{\mu}^\ell= f(\bm{w}^{\ell\pm1}\bm{a}^{\ell\pm1})$ & $\bm{\mu}= f(\bm{w}\bm{a})$     \\
Error      & $\bm{\epsilon}^\ell=\bm{a}^\ell-\bm{\mu}^\ell$         & $\bm{\epsilon}=\bm{a}-\bm{\mu}$ \\
Energy     & $\frac{1}{2}\sum_{\ell}(\bm\epsilon^\ell)^2$           & $\frac{1}{2}\bm\epsilon^2$      \\
Inference &
  $\Delta \bm{a}^\ell= -\gamma\left( \bm{\epsilon}^\ell-\left(\bm{w}^\ell   \right)^T \bm{\epsilon}^{\ell\mp 1}\odot f'\left( \bm{w}^\ell\bm{a}^\ell  \right)\right)$ &
  $\Delta \bm{a} = - \gamma \left( \bm{\epsilon}-\bm{w} ^T   \bm{\epsilon}\odot f'\left( \bm{w} \bm{a} \right) \right)$ \\
Learning &
  $\Delta \bm{w}^\ell = \alpha \bm{\epsilon}^{l\mp1}\odot   f'\left(\bm{w}^\ell\hat{\bm{a}}^\ell\right)(\hat{\bm{a}}^\ell)^T$ &
  $\Delta \bm{w} = \alpha\bm{\epsilon}\odot   f'\left(\bm{w}\hat{\bm{a}}\right)\hat{\bm{a}}^T$ \\ \hline
\end{tabular}
\label{tab:PCG_Vs_PCN}
\end{table}



Algorithm \ref{alg:ZIL} shows pseudocode for Z-IL, cf. section \ref{sec:BP-relation} in the main text. Note that to write down Z-IL most simply we here use notation that contradicts other sections. We use $\bm \mu^\ell = f(\bm w^{\ell+1}\bm a^{\ell+1})$, with data $\bm x^{(n)}$ clamped to $\bm a^L$ to and $\bm y^{(n)}$ to $\bm a^0$.
\begin{algorithm}
\begin{algorithmic}[1]
\REQUIRE: $\bm \mu = f(\bm w^{\ell+1}\bm a^{\ell+1})$ \COMMENT{Downward predictions}
\REQUIRE: $\bm a^L = \bm x^{(n)}$, $\bm a^0 = \bm y^{(n)}$  \COMMENT{Clamp data}
\REQUIRE: $\bm a^\ell(0)=\bm \mu^\ell(0)$ for $\ell =1, \ldots, L-1$ \COMMENT{Feedforward-pass initialization}
\REQUIRE: $\gamma=1$ 
\FOR {$t=0$ to $L$} \COMMENT{Note: $T=L$}
\FOR{each $\ell$} 
\STATE $\bm{a}^\ell(t+1)= \bm{a}^\ell(t)-\gamma\pdv{E}{\bm{a}^\ell}$ 
\STATE $\bm{w}^{\ell=t}(t+1)= \bm{w}^{\ell=t}(t)-\alpha\pdv{E}{\bm{w}^{\ell=t}}$ \COMMENT{Note: $\ell=t$}
\ENDFOR
\ENDFOR
\end{algorithmic}
\caption{Learning $\{\bm x^{(n)}, \bm y^{(n}\}$ with Z-IL, with updates equal to BP with MSE loss.\label{alg:ZIL}}
\end{algorithm}

\section{Computational Complexity}\label{sec:app_CC}
Here, we compare the time complexity of BP and IL. The practical computation time of these algorithms consists of (1) the complexity of one weight update, and (2) the number of weight updates until convergence, but the latter cannot be computed in general, since it depends on factors such as the dataset and optimizer used. However, (1) is tractable by considering the calculations in table \ref{tab:comp_BP_IL}. The most costly computations for each line are the matrix multiplications (including outer products). We can use the following:
\begin{itemize}
\item A matrix multiplication $(m\times n)\times(n\times p)$ has time complexity $\mathcal{O}(mnp)$.
\item The outer product of $m$-dim vector with $n$-dim vector is a matrix multiplication $(m\times 1)\times(1\times n)$.
\end{itemize}
For example, this means that the matrix multiplication $\bm w^{\ell-1}\bm a^{\ell-1}$ has a complexity  of $\mathcal{O}(n^\ell n^{\ell+1})$. For convenience we define the complexity of the largest matrix multiplication as:
\begin{equation}
M = \max(\{ n^\ell n^{\ell+1}\}_{\ell})~.
\end{equation}

The results are shown in tables \ref{tab:CC_BP} for BP and \ref{tab:CC_IL} for IL and incremental IL. For both, the complexity without parallelization is shown, along with the results for parallelized layers ([PL]). It can be seen that with parallelized layers, standard IL is faster only if $T<L$, whereas incremental IL is faster by a factor $L$. For incremental IL, it should be noted that although the time complexity \textit{per weight update} no longer scales with $T$, implementations so far suggest that, for a given epoch, one should still spend several inference iterations (i.e. E-steps) on a given datapoint/batch (in contrast with BP, which always performs a single forward and backward pass per batch). 
This means the \textit{total} complexity would increase again with this factor. At the same time, it has been shown that low values $T$ (e.g., $T=3$ or $5$ \cite{salvatori_stable_2024, alonso_understanding_2024}) appear to be sufficient, implying a substantial speed-up when $L$ is large (which is common in modern deep learning). However, it is still unknown what values of $T$ are appropriate in general.
\begin{table}[h!]
\centering
\setcellgapes{4pt} 
\makegapedcells
\begin{tabular}{lllll}
\hline
\textbf{Phase}          & \textbf{TC/layer} & \textbf{Layers}                  & \textbf{BP}   & \textbf{BP [PL]} \\ \hline
Forward pass   & $M$      & $\ell : 0\rightarrow L$ & $LM$ & $LM$    \\
Backward pass  & $M$      & $\ell : L\rightarrow 1$ & $LM$ & $LM$    \\
Weight update  & $M$      & $\forall \ell$          & $LM$ & $M$     \\ \hline
\textbf{Total} &          &                         & $LM$ & $LM$    \\ \hline
\end{tabular}
\vspace{1em}
\caption{\textbf{Time complexity (TC) of a single weight update for BP} in an FNN. TC is quantified with big-O notation where constant factors are ignored. [PL] refers to TC with parallelized layers, which has not yet been achieved in practice.\label{tab:CC_BP}}
\end{table}
\begin{table}[h!]
\centering
\setcellgapes{4pt} 
\makegapedcells
\begin{tabular}{lllllll}
\hline
\textbf{Phase}     & \textbf{TC/layer/iteration} & \textbf{Time step, layers}                  & \textbf{IL}    & \textbf{IL [PL]} &\textbf{ i-IL} & \textbf{i-IL [PL]} \\ \hline
Inference & $M$                & $t:0\rightarrow T,\, \forall \ell$ & $TLM$ & $TM$    & $LM$           & $M$                 \\
Learning & $M$ & $\forall \ell$ & $LM$  & $M$  & $LM$ & $M$ \\ \hline
 \textbf{Total}        &     &                & $TLM$ & $TM$ & $LM$ & $M$ \\ \hline
\end{tabular}
\vspace{1em}
\caption{\textbf{Time complexity (TC) of a single weight update for IL and incremental IL (i-IL)} in a PCN. TC is quantified with big-O notation where constant factors are ignored. [PL] refers to TC with parallelized layers, which has not yet been achieved in practice.\label{tab:CC_IL}}
\end{table}

\end{document}